\documentclass[10pt, a4paper]{article}
\usepackage[top=3cm, bottom=4cm, left=3.5cm, right=3.5cm]{geometry}

\usepackage{amsmath,amsthm,amsfonts,amssymb,amscd}
\usepackage{fancyhdr}
\usepackage{graphicx}
\usepackage{float}
\usepackage{color, comment}
\usepackage{environ}
\usepackage{enumerate}
\usepackage{subcaption}
\usepackage[shortlabels]{enumitem}
\usepackage{sectsty}
\usepackage{listings}
\usepackage{thmtools}
\usepackage{caption}

\usepackage{color,soul}
\usepackage{shadethm}
\usepackage{hyperref}
\usepackage{setspace}
\usepackage[linguistics]{forest}

\hypersetup{
    colorlinks=true,
    linkcolor=blue,
    filecolor=magenta,      
    urlcolor=blue,
}

\catcode`\<=\active \def<{
\fontencoding{T1}\selectfont\symbol{60}\fontencoding{\encodingdefault}}
\catcode`\>=\active \def>{
\fontencoding{T1}\selectfont\symbol{62}\fontencoding{\encodingdefault}}
\catcode`\<=\active \def<{
\fontencoding{T1}\selectfont\symbol{60}\fontencoding{\encodingdefault}}

\begin{document}

\begin{titlepage}
    \begin{center}
        \vspace*{1.5cm}
            
        \LARGE
        \textbf{Iterative Magnitude Pruning as a Renormalisation Group: A Study in The Context of The Lottery Ticket Hypothesis }
            
        \vspace{2cm}
        \large
        \textbf{Abu-Al Hassan}\\
            
        \vspace{1.5cm}
        
        \large

        \vfill
            
        \vspace{1cm}
        \large
        
        \today
            
    \end{center}
\end{titlepage}

\setlength{\parindent}{0pt}

{
\centering
\section*{Abstract}
}

This thesis delves into the intricate world of Deep Neural Networks (DNNs), focusing on the exciting concept of the Lottery Ticket Hypothesis (LTH). The LTH posits that within extensive DNNs, smaller, trainable subnetworks — termed "winning tickets" — can achieve performance comparable to the full model. A key process in LTH, Iterative Magnitude Pruning (IMP), incrementally eliminates minimal weights, emulating stepwise learning in DNNs. Once we identify these winning tickets, we further investigate their "universality" - that is, we check if a winning ticket that works well for one specific problem could also work well for other, similar problems.  We also bridge the divide between the IMP and the Renormalisation Group (RG) theory in physics, promoting a more rigorous understanding of IMP.\\

\newpage
\tableofcontents

\newpage

\section{Introduction}
Deep Neural Networks (DNNs), despite their impressive capabilities, often entail a considerable computational overhead due to the sheer magnitude of parameters — typically between $10^{6}$ to $10^{11}$ \cite{DNN} \cite{LeCun2015}. This vastness tends to decelerate the training process. One potent strategy to combat this computational bottleneck is pruning — eliminating superfluous connections or neurons, which reduces the number of computationally expensive parameters, thereby accelerating prediction times.\\

Our thesis is devoted to an in-depth exploration of the Lottery Ticket Hypothesis (LTH), a groundbreaking idea in the realm of deep learning \cite{LTH} \cite{morcos2019lotteryscale} \cite{zhou2019deconstructing}. According to LTH, there exist "winning tickets" — smaller subnetworks embedded within DNNs — that can be trained to match or even outperform the full models. This thesis seeks to uncover these winning tickets and test their universality, i.e., a winning ticket that is successful for one task (the specific problem a neural network is designed to solve) may also be efficacious for other tasks within the same class (similar problem types) \cite{redman} \cite{COMPUTERVISION} \cite{zhou2019deconstructing}. This principle could potentially expedite training times and expedite the development of high-performing models across a range of tasks within the same class. \\ 

An integral component of our investigation is the technique of Iterative Magnitude Pruning (IMP) \cite{IMP} \cite{IMP2}, which facilitates the discovery of winning tickets by gradually eliminating the least significant weights. In a more theoretical vein, our thesis draws connections between IMP and the Renormalisation Group (RG) theory, a powerful mathematical framework in physics \cite{GOLDENFELD}. RG theory provides insights into how transformations in a physical system unfold at various scales. Similarly, in IMP, we view different parameter densities as variations in scale. We show that IMP is an RG scheme \cite{redman}. By applying methodologies from the Renormalisation Group theory to IMP, we aspire to foster a more rigorous and generalisable understanding of IMP, which is currently bereft of an effective theory \cite[p.~1]{NO_IMP_THEORY}. This endeavour could potentially amplify the efficacy and reliability of IMP and, in turn, substantially impact the field of deep learning.

\section{Introduction to Iterative Magnitude Pruning} 
Iterative Magnitude Pruning (IMP) initiates by fully training a neural network and subsequently discarding a fraction of the least significant weights. Following this pruning step, the residual weights revert to their initial configuration. This cycle of training, pruning, and resetting transpires iteratively, progressively unearthing an efficient subnetwork, or "winning ticket". This methodology, grounded in the Lottery Ticket Hypothesis, provides a systematic means to discover these winning tickets within larger, often unwieldy neural networks. \\

While IMP is favoured in practice due to its conceptual simplicity, ease of implementation, and efficacy, theoretical explanations of its effectiveness have been limited \cite{balwani2022zeroth}. Balwani and Krzyston's study \cite{balwani2022zeroth} shows that IMP preferentially retains weights that maintain network topology, providing unique insights into the extent of pruning possible without affecting zeroth order topological features.\\

\hrule
\vspace{10pt}
\begin{equation}
\text{{\textbf{Algorithm: }}} \text{{Iterative Magnitude Pruning }} 
\label{eq:IMPALGO}
\end{equation}
\vspace{10pt}
\hrule
\begin{flalign*}
& \text { \textbf{Input:} Loss function } L: \mathbb{R}^p \rightarrow \mathbb{R} \text {, training time } T \in \mathbb{R}_{+}, \text {initialisation } \boldsymbol{w}^{\text {init }} \in \mathbb{R}^p \text {, iterations } \\
& \text { of pruning } q<p . \\
& \text { \textbf{Output:} } \boldsymbol{w}^{(q)}(T) &
\end{flalign*}
\hrule

\begin{flalign*}
& \textbf{Method for IMP at x\% level:} \hspace{7cm} \text{ (adaptation of \cite[p.~2]{IMP})}   \\
& \text{Set } M^{0}=\mathbb{I}_p \\
& \text{for } k=0 \text{ to } q \text{ do: } \\
& \begin{aligned}
& \quad \text{Initialise } \boldsymbol{w}^{(k)}(0)=M^{k} \boldsymbol{w}^{\text{init}} \\
& \quad \text{Train } \dot{\boldsymbol{w}}^{(k)}(t)=-M^{k} \nabla L\left(\boldsymbol{w}^{(k)}(t)\right) \text{ for } t \in[0, T] \\
& \quad \text{Prune the smallest } x\% \text{ of } \left\{\left|w_j^{(k)}(T)\right|: M^{k}_{j j}=1\right\} \text{ and set corresponding } M^{k}_{i i}=0 \\
& \quad \text{Set } M^{k+1}=M^{k}
\end{aligned} \\
& \text{return } \boldsymbol{w}^{(q)}(T) &
\end{flalign*}
\hrule
\vspace{5pt}

Where $\boldsymbol{w}^{\text{init }}\in\mathbb{R}^p$ is the initial set of weights for the neural network. \\

$q$ represents the number of pruning steps. It must be less than the number of parameters in the network, $p$. \\

$M$ is called the mask as the diagonal entries of this matrix will indicate which weights are active (1) and which are pruned (0). \\

Train $\dot{\boldsymbol{w}}^{(k)}(t)=-M \nabla L\left(\boldsymbol{w}^{(k)}(t)\right)$ for $t \in[0, T]$: This line trains the network by following the negative gradient of the loss function, with the modification that pruned weights don't get updated. \\

While technically, biases in a neural network are also considered weights, it's important to understand their unique role in model learning and performance. Biases are employed to shift the activation function to the left or right, which can prove crucial in successful learning and adaptation to data. Pruning biases might, therefore, impose a greater negative impact on a network's performance than pruning weights. \\

Moreover, the count of bias terms in a neural network is usually significantly less than that of weights. Therefore, retaining biases does not substantially contribute to model complexity. This consideration, coupled with the potential for enhanced accuracy, justifies the exclusion of biases from the pruning process. This approach is consistent with the methodology proposed in the original Lottery Ticket Hypothesis paper by Frankle and Carbin \cite{LTH}.\\

For a theoretical understanding of neural network pruning and the effects of pruning on the properties and capabilities of the neural network, you can read \cite{yang2023neuraltangent,dogred20}.

\section{Lottery Ticket Hypothesis}
The Lottery Ticket Hypothesis (LTH), an intriguing concept in the field of deep learning, was proposed by Frankle and Carbin in 2019 \cite{LTH}. This hypothesis posits that randomly initialized, dense neural networks embed subnetworks – known as "winning tickets". These subnetworks, when trained in isolation, are capable of achieving comparable accuracy to the original, dense network, but with less computational time. Winning tickets are identified using the previously discussed method of Iterative Magnitude Pruning (see section 2). The LTH derives its name from the analogy of unearthing these high-performing subnetworks within the complex web of a dense network as akin to finding a winning ticket in a lottery. By facilitating the discovery of these "winning tickets", the LTH offers a strategy to drastically reduce the computational resources required for training DNNs, without diminishing their performance – thereby boosting efficiency.\\

The size of a winning ticket, or the pruned network, hinges upon several factors: the specific task at hand, the model's architectural design, the optimization algorithm, and the pruning strategy employed – whether Magnitude-based Pruning, Sensitivity-based Pruning, or Random Pruning etc \cite{ARCHITECTURE} \cite{PRUNE_FILTERS} \cite{LTH}. However, there is no strict size constraint that defines a subnetwork as a winning ticket. Empirical evidence from experiments conducted by Frankle and Carbin in 2019 suggests that winning tickets containing less than 10-20\% of the parameters of the original network can operate without sacrificing accuracy \cite[p.~1]{LTH}.

\subsection{Identifying winning tickets from Iterative Magnitude Pruning}
At each iteration of IMP, a subnetwork is derived. Assume we execute $q$ iterations of IMP with initial weights denoted as $\boldsymbol{w}^{\text {init }}$, following the procedure outlined in \eqref{eq:IMPALGO}. We conclude with a subnetwork characterized by weights $\boldsymbol{w}^{(q)}(T)$ and its corresponding mask matrix $M^{q}$. Subsequently, we initialize the same neural network with weights $\boldsymbol {w}=M^{q} \boldsymbol{w}^{\text {init }}$. If the model, when trained from $\boldsymbol {w}$, attains an accuracy comparable to the full model or better (model trained from $\boldsymbol{w}^{\text {init }}$), we classify $\boldsymbol {w}=M^{q} \boldsymbol{w}^{\text {init }}$ as a winning ticket for the model. \\

\subsection{Universality of winning tickets}
The Lottery Ticket Hypothesis posits that winning tickets $\boldsymbol{w}$ are task-specific, suggesting that a winning ticket identified for one task might not exhibit high performance on a different task \cite{LTH}. The inherent structure and connections of a winning ticket are optimized for a specific task and may not generalize well to others.\\

Nonetheless, recent studies have unearthed evidence that winning tickets can exhibit transferability to related tasks \cite{redman} \cite{COMPUTERVISION} \cite{Sabatelli2020}. For instance, if two tasks share close relations, a winning ticket $\boldsymbol{w}$ discovered for one task might still exhibit commendable performance on the other task, albeit not optimal.\\

Additionally, research \cite{ELTH} demonstrates that winning tickets can be transferred between disparate architectures. This signifies the capability to utilize a winning ticket identified in one neural network architecture as an initialization point for training in a different architecture.\\

These observations imply the feasibility of employing winning tickets to investigate the similarities between “tasks” and “architectures”.

\section{Renormalisation Group theory}
The central concept in Renormalisation Group theory is the concept of "scaling." Many systems exhibit behaviour that is "scale invariant," meaning that their properties remain the same under a change in scale \cite{INVARIANT1} \cite{INVARIANT2}. In such systems, physical quantities often follow a "power law" behaviour, with quantities scaling as some power of the scale factor.\\

The Renormalisation Group theory is particularly useful for understanding the behaviour of systems near "critical points," where the system undergoes a phase transition. For instance, near the critical temperature, many systems exhibit power-law scaling behaviour. In such cases, the power-law exponent, known as the "critical exponent," can provide significant insights into the behaviour of the system.\\

The theory also provides a way to classify systems into "universality classes." Systems within the same universality class have the same critical exponents, indicating that they behave in the same way near their critical points, regardless of the specifics of their microscopic interactions. This has significant implications for the study of complex systems, as it allows for the prediction of macroscopic behaviour from a limited understanding of microscopic interactions.

\subsection{Block Spins argument by L.P. Kadanoff }
The 'Block Spins' argument introduced by Leo P. Kadanoff \cite{kadanoff2000statics} is one of the seminal concepts that formed the basis of modern Renormalization Group (RG) theory. By introducing the concept of block spins and coarse-graining, Kadanoff established a connection between microscopic details and macroscopic behaviour. His approach led to the development of the renormalization group theory. Kandanoff introduces the concept of scaling. The idea of scaling arises from the observation that, near the critical point, systems exhibit similar behaviour on different length scales. Kadanoff's approach to coarse-graining and block spins helps to demonstrate that the free energy of the original system and the block spin system have the same functional form, with the block spin system effectively scaling the original system.\\

This scaling behaviour leads to the development of scaling laws and critical exponents, which are crucial for understanding the properties of systems near critical points.\\

Consider a d-dimensional grid with spacing $a$, where the system has the Hamiltonian function (energy function)\cite[pp.~230--235]{GOLDENFELD}:\\

\[
\begin{aligned}
\beta H_{\Omega} & =-\beta J \sum_{\langle i j\rangle=1}^{N} S_{i} S_{j}-\beta H \sum_{i} S_{i} \\
& \equiv-K \sum_{\langle i j\rangle} S_{i} S_{j}-h \sum_{i=1}^{N} S_{i}
\end{aligned}
\]

with

\[
\begin{aligned}
K & \equiv \beta J \\
h & =\beta H .
\end{aligned}
\]

On the grid, we can replace a block of side $\ell  a$ that contains $\ell^{d}$ with a single 'block spin'. Then the total number of block spins is $N  \ell^{-d}$ for some $N$.\\

We can define the spin of block I to be $S_{I}$:
\[
S_{I} \equiv \frac{1}{\left|\overline{m_{\ell}}\right|} \frac{1}{\ell^{d}} \sum_{i \in I} S_{i}
\]

Where $\bar{m}_{\ell}$ is the average magnetization of block we defined as:

\[
\bar{m}_{\ell} \equiv \frac{1}{\ell^{d}} \sum_{i \in I}\left\langle S_{i}\right\rangle
\]

After this normalisation the magnitudes of the spins are the same as the original system (non-coarsed):
\[
\left\langle S_{I}\right\rangle= \pm 1
\]

Block spin renormalisation theory has two assumptions:\\
\textbf{The first assumption} is that since in the original system, spins interact with only nearest-neighbour spins and the external field, we can assume that the new blocks also interact with the nearest neighbour block spins and an effective external field.\\

As a result of this assumption, we need to define new coupling constants between the block spins and the effective external field. We can write these as $K_{\ell}$ and $h_{\ell}$, where for the original system $\ell=1$. With this the Hamiltonian becomes:
\[
-\beta H_{\ell}=K_{\ell} \sum_{\langle I J\rangle}^{N \ell^{-d}} S_{I} S_{J}+h_{\ell} \sum_{I=1}^{N \ell^{-d}} S_{I},
\]

Notice this has the same form as the original Hamiltonian except. This new system has fewer spins than the original system.\\

The system with Hamiltonian $H_{\ell}$ is further away from criticality than the original system $H_{\Omega}$. It also has a new effective reduced temperature, $t_{\ell}$ and an effective magnetization field $h_{\ell}$. The reduced temperature, t, measures how far the temperature of the system is from the critical temperature ($T_c$).\\

\[
t \equiv \frac{T-T_{c}}{T_{c}}
\]

The relationship between effective reduced temperature and effective magnetization field is:
\[
h_{\ell}=h \bar{m}_{\ell} l^{d} \text {. }
\]

Since $H_{\ell}$ has the same form as $H_{\Omega}$, which means that the free energy of the block spin system will also have a similar form as the original spin system but with $t_\ell$ and $h_\ell$ instead of $t$ and $h$.

The free energy per spin (or block spin) of the original system is related to that of the block spin system by:
\[
N  \ell^{-d}  f_s(t_\ell, h_\ell) = N  f_s(t, h)
\]

Leading to the functional form of the free energy per spin changes under the block spin transformation:
\[
f_s(t_\ell, h_\ell) = \ell^{d} * f_s(t, h)
\]

\textbf{The second assumption} is about how the reduced temperature ($t_\ell$) and external field ($h_\ell$) change during the transformation. We assume that:
\[
\begin{aligned}
t_{\ell} & =t \ell^{y_{t}} \quad y_{t}>0 \\
h_{\ell} & =h \ell^{y_{h}} \quad y_{h}>0 .
\end{aligned}
\]

They are both dependent on the block size $\ell$ and we don't know $y_{t}$ and $y_{h}$ yet but  assume they are positive.\\

Now we can write the relationship between the free energy per spin of the original system and the block spin system:

\begin{equation}
f_{s}(t, h)=\ell^{-d} f_{s}\left(t \ell^{y_{t}}, h \ell^{y_{h}}\right)
\label{eq:scaling_relation}
\end{equation}

Kadanoff's block spin argument helps us understand the form of scaling relations, but it doesn't give us the exponents $y_t$ and $y_h$. While Kadanoff's block spin idea provided a heuristic way to understand how systems change under coarse-graining, it didn't offer a precise, quantitative method for predicting these changes.\\

\subsection{Critical Phenomena and Renormalisation Group}
Having introduced 'Block spin' argument by Kadanoff, we build on this by introducing Wilson's RG approach \cite{wilson1975renormalization}. Wilson extended the idea of block spins to field theories, providing a mathematical framework that is applicable to a broad range of systems beyond Ising-type spin models. He answered the question of how and why systems' properties change under transformations of scale and provides a way to understand the "fixed points" of this flow, which correspond to scale-invariant phases of the system. Kadanoff's original block spin argument didn't include these key concepts.\\

We will now detail coarse-graining transformation by examining the characteristics of the block spin transformations \cite[pp.~236--239]{GOLDENFELD}. The key idea is that after performing a block spin transformation, the distance between block spins is a. If we rescale the lengths so that the new distance between block spins is the same as the original distance between microscopic spins, the system appears similar to the original system but with a different Hamiltonian. Repeating these steps produces a series of Hamiltonians, each describing systems that are further from criticality.\\

Let's consider a Hamiltonian described as:
\[
\mathcal{H} \equiv-\beta H_{\Omega}=\sum_{n} K_{n} \Theta_{n}\{S\}
\]

Here, $K_n$ are the coupling constants, and $\Theta_n{S}$ are the local operators, which are functionals of the degrees of freedom ${S}$.

I will now describe how the coupling constants change when we "zoom out" and look at the system at a larger scale.

The equation:
\[
\left[K^{\prime}\right] \equiv R_{\ell}[K] \quad \ell>1
\]
tells us that when we apply the RGT, the original set of coupling constants $K$ transforms into a new set $K^{\prime}$.\\

To calculate the RGT, we first define the partition function $Z_{N}[K]$, which is a mathematical tool used to analyze the statistical properties of a system, and a quantity $g[K]$ that is related to the free energy per degree of freedom:
$$
Z_{N}[K]=\operatorname{Tr} e^{\mathcal{H}}
$$

$$
g[K] \equiv \frac{1}{N} \log Z_{N}[K]
$$

The partition function is crucial because many thermodynamic properties, such as internal energy, entropy, and free energy, can be derived from it. It is a measure that encodes the statistical properties of a system in equilibrium.\\

The RGT reduces degrees of freedom by $\ell^d$, creating a new effective Hamiltonian for "block variables" ${S_{I}^{\prime}}$ by taking a partial trace over original degrees of freedom ${S_{i}}$, while block degrees of freedom are fixed.

$$
\begin{aligned}
e^{\mathcal{H}_{N}^{\prime}\left\{\left[K^{\prime}\right], S_{I}^{\prime}\right\}} & =\operatorname{Tr}_{\left\{S_{i}\right\}} e^{\mathcal{H}_{N}\left\{[K], S_{i}\right\}} \\
& =\operatorname{Tr}\left\{S_{i}\right\} P\left(S_{i}, S_{I}^{\prime}\right) e^{\mathcal{H}_{N}\left\{[K], S_{i}\right\}}
\end{aligned}
$$

$P\left(S_{i}, S_{I}^{\prime}\right)$ is a projection operator used to allow unrestricted trace in the equation. It ensures that the range of values for coarse-grained degrees of freedom $S_{I}^{\prime}$ matches that of the original degrees of freedom $S_{i}$.\\

Let's work with the Ising spins on a square lattice. We define an RGT transformation using blocks of linear dimension $(2\ell+1)a$, where a is the lattice spacing. The block spin $S_I'$ is given by:

$$
S_{I}^{\prime}=\operatorname{sign}\left(\sum_{i \in I} S_{i}\right)= \pm 1 .
$$

This means we assign the block spin value based on the sign of the sum of spins within the block. The associated projection operator is defined as:
$$
P\left(S_{i}, S_{I}^{\prime}\right)=\prod_{I} \delta\left(S_{I}^{\prime}-\operatorname{sign}\left[\sum_{i \in I} S_{i}\right]\right)
$$

The projection operator must satisfy three requirements:\\
(i) $P\left(S_{i}, S_{I}^{\prime}\right) \geq 0$\\
(ii) $P\left(S_{i}, S_{1}^{\prime}\right)$ reflects the symmetries of the system;\\
(iii) $\sum_{\left\{S_{1}^{\prime}\right\}} P\left(S_{i}, S_{I}^{\prime}\right)=1$ \\

Condition (i) ensures that the exponential term of the transformed Hamiltonian is non-negative, allowing us to identify the effective Hamiltonian for the new degrees of freedom, $S_{I}^{\prime}$.\\

In condition (ii) by symmetry, it is meant that the proposed operator does not introduce any new or unauthorised couplings that were not possible in the original, non-coarse-grained system.

For example, if the original Hamiltonian has the form:
$$
\mathcal{H}_{N}=N K_{0}+h \sum_{i} S_{i}+K_{1} \sum_{i j} S_{i} S_{j}+K_{2} \sum_{i j k} S_{i} S_{j} S_{k}+\ldots
$$
The transformed Hamiltonian will have the same form, but with new, transformed coupling constants:
$$
\mathcal{H}_{N^{\prime}}^{\prime}=N^{\prime} K_{0}^{\prime}+h^{\prime} \sum_{I} S_{I}^{\prime}+K_{1}^{\prime} \sum_{I J} S_{I}^{\prime} S_{J}^{\prime}+K_{2}^{\prime} \sum_{I J K} S_{I}^{\prime} S_{J}^{\prime} S_{K}^{\prime}+\ldots
$$

Condition (iii) ensures a well-defined projection operator that has a clear, one-to-one mapping between the original and new parameters, and if probabilistic, the sum of all probabilities equals 1. Probabilistic Renormalisation Group operators are often used to study complex or disordered systems where deterministic Renormalisation Group operators are not well-suited. \\

As a result, the partition function is invariant under the Renormalisation Group transformation thus preserving the statistical properties and thermodynamic behaviour of the original system.\\

$$
\begin{aligned}
& Z_{N^{\prime}}\left[K^{\prime}\right] \equiv \operatorname{Tr}\left\{S_{I}^{\prime}\right\} e^{\mathcal{H}^{\prime}{ }_{N}\left\{\left\{K^{\prime}\right], S_{I}^{\prime}\right\}} \\
& =\operatorname{Tr}\left\{S_{I}^{\prime}\right\} \operatorname{Tr}\left\{S_{I}\right\} P\left(S_{i}, S_{I}^{\prime}\right) e^{\mathcal{H}_{N}\left\{[K], S_{i}\right\}} \\
& =\operatorname{Tr}\left\{S_{i}\right\} e^{\mathcal{H}_{N}\left\{[K], S_{i}\right\}} \cdot 1 \\
& =Z_{N}[K]
\end{aligned}
$$

Which gives,
$$
\begin{aligned}
\frac{1}{N} \log Z_{N}[K] & =\frac{\ell^{d}}{\ell^{d} N} \log Z_{N^{\prime}}\left[K^{\prime}\right] \\
& =\ell^{-d} \frac{1}{N^{\prime}} \log Z_{N^{\prime}}\left[K^{\prime}\right]
\end{aligned}
$$

leading to,
$$
g[K]=\ell^{-d} g\left[K^{\prime}\right]
$$

Thus the free energy per degree of freedom is related to the transformed free energy by a factor of $\ell^{-d}$.

\subsection[Fixed points]{Fixed points\cite[pp.~242--246]{GOLDENFELD}}
In order to understand the behaviour of systems undergoing repeated applications of renormalization group (RG) transformations, we observe the evolution or "flow" of parameters. These parameters, originating from a wide range of initial values, constitute what is termed the renormalization group flow.\\

Interestingly, an aspect of this flow is its common tendency to gravitate towards specific points, known as fixed points. Fixed points in the space of coupling constants by definition don't change under Renormalisation Group transformations. In the vicinity of these fixed points, systems tend to exhibit a characteristic scaling behaviour. This behaviour implies that these systems manifest consistent patterns or characteristics, irrespective of their scale or size, showcasing the universality of such patterns. \\

Mathematically a fixed point is represented by:
$$
\left[K^{*}\right]=R_{\ell}\left[K^{*}\right]
$$

Let's consider a Hamiltonian close to the fixed point Hamiltonian. We write it as:
$$\mathcal{H}=\mathcal{H}\left[K^{*}\right] \equiv \mathcal{H}^{*}$$ 
$$\mathcal{H}=\mathcal{H}^{*}+\delta \mathcal{H}$$.

After performing an Renormalisation Group transformation: $\left[K^{\prime}\right]=R_{\ell}[K]$, the new coupling constants, denoted by $K_n'$, can be written as:

$$
K_{n}^{\prime}=K_{n}^{\prime}[K] \equiv K_{n}^{*}+\delta K_{n}^{\prime}
$$

By Taylors theorem,
$$
K_{n}^{\prime}\left\{K_{1}^{*}+\delta K_{1}, K_{2}^{*}+\delta K_{2}, \ldots\right\}=K_{n}^{*}+\left.\sum_{m} \frac{\partial K_{n}^{\prime}}{\partial K_{m}}\right|_{K_{m}=K_{m}^{*}} \cdot \delta K_{m}+O\left(\left(\delta K^{\prime}\right)\right.$$

Allowing us to express the change in coupling constants after the transformation, $\delta K_n'$, in terms of the original changes, $\delta K_n$.
$$
\delta K_{m}^{\prime}=\sum_{m} M_{n m} \delta K_{m}
$$

Where $M_{n m}$ is the partial derivative of the new coupling constants with respect to the original ones, evaluated at the fixed point values. The matrix $M$ is the linearized Renormalisation Group transformation near the fixed point. For simplicity, we can assume that $M$ is symmetric.\\

To examine Renormalisation Group flows near the fixed point, we employ the linearised Renormalisation Group transformation denoted by $M^{(\ell)}$ and then study the eigenvalues and eigenvectors. "Flow" refers to the evolving system properties and interactions between regimes as the scale changes.\\

The eigenvalues and eigenvectors of $M^{(\ell)}$ are denoted as $\Lambda_{\ell}^{(\sigma)}$ and $e_{n}^{(\sigma)}$, where $\sigma$ identifies the eigenvalues and $n$ refers to the vector components. Employing the Einstein summation convention, we get:

$$
M_{n m}^{(\ell)} e_{m}^{(\sigma)}=\Lambda^{(\sigma)} e_{n}^{(\sigma)}
$$

Using the associativity of matrices,
$$
\mathbf{M}^{(\ell)} \mathbf{M}^{\left(\ell^{\prime}\right)}=\mathbf{M}^{\left(\ell \ell^{\prime}\right)}
$$
giving us,
$$
\Lambda_{\ell}^{(\sigma)} \Lambda_{\ell^{\prime}}^{(\sigma)}=\Lambda_{\ell \ell^{\prime}}^{(\sigma)}
$$

By differentiating with respect to $\ell^{\prime}$, setting $\ell^{\prime}=1$, and solving the obtained differential equation, we derive:
$$
\Lambda_{(\ell)}^{(\sigma)}=\ell^{y_{\sigma}}
$$

Here, $y_{\sigma}$ is a number to be determined, but it is independent of $\ell$.\\

This shows that the eigenvalues can be expressed as a power of the scale factor, $\ell$. This information can help us understand how the system behaves near fixed points as the scale changes.\\

I will now explore how the changes in coupling constants, denoted by $\delta K$, transform under the linearized Renormalisation Group transformation $M$. We can express $\delta K$ in terms of the eigenvectors of $M$:
$$
\delta \mathbf{K}=\sum_{\sigma} a^{(\sigma)} \mathbf{e}^{(\sigma)}
$$

In this case, we express $[K]$ as a vector $\mathrm{K}=\left(K_{1}, K_{2}, \ldots\right)$. The orthonormality of eigenvectors is assumed to determine the coefficients $a^{(\sigma)}$:

When we apply the linearized Renormalisation Group transformation $\mathbf{M}$, we get:

$$
\begin{aligned}
\delta \mathbf{K}^{\prime} & =\mathbf{M} \delta \mathbf{K} \\
& =\mathbf{M} \sum_{\sigma} a^{(\sigma)} \mathbf{e}^{(\sigma)} \\
& =\sum_{\sigma} a^{(\sigma)} \Lambda^{(\sigma)} \mathbf{e}^{(\sigma)} \equiv \sum_{\sigma} a^{(\sigma) \prime} \mathbf{e}^{(\sigma)},
\end{aligned}
$$

Here, we define $a^{(\sigma) \prime}$ as the projection of $\delta K^{\prime}$ in the direction $\mathrm{e}^{(\sigma)}$. This equation is crucial because it tells us that some components of $\delta \mathrm{K}$ grow under $M^{(\ell)}$ while others shrink. If we arrange the eigenvalues by their absolute value,\\

The eigenvalue absolute values follow the order:
$$
\left|\Lambda_{1}\right| \geq\left|\Lambda_{2}\right| \geq\left|\Lambda_{3}\right|
$$
We consider three cases:
\begin{enumerate}
\item $\left|\Lambda^{(\sigma)}\right|>1$ or $y^{\sigma}>0$: $a^{(\sigma) \prime}$ grows with increasing $\ell$.
\item $\left|\Lambda^{(\sigma)}\right|<1$ or $y^{\sigma}<0$: $a^{(\sigma)}$ diminishes with increasing $\ell$.
\item $\left|\Lambda^{(\sigma)}\right|=1$ or $y^{\sigma}=0$: $a^{(\sigma) \prime}$ remains constant as $\ell$ increases.
\end{enumerate}

After applying $\mathrm{M}^{(\ell)}$ multiple times, only components of $\delta K$ along directions $\mathrm{e}^{(\sigma)}$ for the first case (a) are significant. Projections of $\delta \mathbf{K}$ in other directions will either shrink or stay fixed. These cases are named:
\begin{enumerate}
\item   Relevant eigenvalues/directions/eigenvectors.
\item   Irrelevant eigenvalues/directions/eigenvectors.
\item   Marginal eigenvalues/directions/eigenvectors.
\end{enumerate}

\subsection{Universality in Renormalisation Group theory}
The concept of universality signifies that disparate systems, despite possessing distinct microscopic details, may exhibit analogous behaviour near critical points--points at which a system undergoes a phase transition. Systems that demonstrate such behaviour are classified into the same 'universality class.' \\

The 'universality class' of a system is dictated by the count and the nature of its pertinent parameters (or directions). Various microscopic models may possess unique parameter coupling constants corresponding to a given critical phenomenon. However, only a handful of these parameter combinations influence the systems' behaviour around critical points. The parameters associated with these influential directions are deemed as relevant. Systems that share an identical count and nature of relevant parameters at a fixed point will exhibit matching critical exponents, thereby categorizing them into the same universality class \cite[p.~4]{redman}. 

Parameters classified as irrelevant do not influence the universality class, as their impact recedes over large scales or at lower energies, a concept encapsulated by the energy-length duality. Irrelevant directions exhibit a negative critical exponent, denoted as $y$. Consequently, an increase in the scale factor ($\ell$) diminishes the contribution of these operators, given that they are effectively multiplied by $\ell^y$, where $y < 0$ \cite{GOLDENFELD}.\\

\subsection[RG Flow]{RG Flow \cite[pp.~3--4]{redman}}
The Renormalisation Group operator, $\mathcal{R}$, simplifies complex systems by replacing local variables with their composite values. The way to formally study $\mathcal{R}$ is by analyzing its effect on the energy function (Hamiltonian). For classical spin systems, the Hamiltonian has a specific form:

\[\mathcal{H}(\mathbf{s}, \mathbf{k})=-\sum_{i} k_{1} s_{i}-\sum_{\langle i, j\rangle} k_{2} s_{i} s_{j}-\ldots,\] \\

Where $s_{i}$ are the spins, $\langle\cdot, \cdot\rangle$ represents nearest neighbor sites on the lattice, and $k_{i}$ are the strengths of the different coupling constants.\\

The Renormalisation Group operator combines spins, creating a new spin system with different coupling constants. The new system is represented by a new Hamiltonian,
\[\mathcal{R \mathcal { H }}(\mathbf{s}, \mathbf{k})=\mathcal{H}\left(\mathbf{s}^{\prime}, \mathcal{T} \mathbf{k}\right)=\mathcal{H}\left(\mathbf{s}^{\prime}, \mathbf{k}^{\prime}\right),\] \\

Which is derived from the original Hamiltonian, $\mathcal{H}(\mathbf{s}^{\prime}, \mathbf{k})$, using the transformed spins, $\mathbf{s}^{\prime}$, and couplings, $\mathbf{k}^{\prime}$, determined by the operator $\mathcal{T}: \mathbb{R}^{K} \rightarrow \mathbb{R}^{K}$.\\

The Renormalisation Group operator, $\mathcal{R}$, can be used to change the Hamiltonian of a spin system and its set of couplings. When $\mathcal{R}$ is applied repeatedly, it generates a flow in the function space of Hamiltonians and the space of coupling constants, which is referred to as Renormalisation Group flow. The Renormalisation Group flow changes depending on the eigenvectors of the linearised operator $\mathcal{T}$ near fixed points, with the eigenvectors defined as relevant ($\lambda_{i}>1$) or irrelevant($\lambda_{i}<1$) based on their eigenvalues.
\subsection[Power-law scaling in Renormalisation Group theory ]{Power-law scaling in Renormalisation Group theory\cite[pp.~252--253]{GOLDENFELD}}
Having gained an understanding of Renormalisation Group flows, let's examine how the Renormalisation Group explains scaling behaviour. The phenomenon of power-law scaling is commonly observed in the critical phenomena studied using Renormalisation Group such as phase transitions in Physics. Phase transitions occur when changing one control parameter within a range leads to the divergence of another parameter called the order parameter or its derivative.\\

For example, near the critical temperature of a ferromagnetic phase transition (where metal goes from being non-magnetic to magnetic as it is cooled), the magnetization display power-law scaling for t such that $t_{L}<t<t_{C}$. The scaling is characterised by the equation $m \sim\left(t_{C}-t\right)^{-\beta}=\Delta t^{-\beta}$, where m is magnetization, $\Delta t$ represents the difference between the critical temperature and the temperature of the system. Here $\beta$ is called the critical exponent.\\

Working with the Ising model, we have two important factors that can change, which are labelled 't' and 'h'. We start with a formula for the free energy density. This is a measure of the energy in the system that could potentially be used to do work.

\[f(t, h)=\ell^{-d} f\left(t^{\prime}, h^{\prime}\right)\]

The new parameters $t'$ and $h'$ are generated by transformations:
\[\begin{aligned}
T^{\prime} & =R_{\ell}^{T}(T, H) \\
H^{\prime} & =R_{\ell}^{H}(T, H)
\end{aligned}\]

Where $R_{\ell}^{T}$ and $R_{\ell}^{H}$ are coarse-graining functions. We then consider the neighbourhood of a fixed point (T*, H*) where
\[\begin{aligned}
T^{*} & =R_{\ell}^{T}\left(T^{*}, H^{*}\right) \\
H^{*} & =R_{\ell}^{T}\left(T^{*}, H^{*}\right) .
\end{aligned}\]

Next, we consider small deviations from the fixed point (T*, H*), called $\Delta T$ and $\Delta H$:
\[\begin{aligned}
\Delta T & =T-T^{*} \\
\Delta H & =H-H^{*}
\end{aligned}\]

\[\left(\begin{array}{c}
\Delta T^{\prime} \\
\Delta H^{\prime}
\end{array}\right)=\mathbf{M}\left(\begin{array}{c}
\Delta T \\
\Delta H
\end{array}\right)\]

with

\[\mathrm{M}=\left(\begin{array}{ll}
\partial R_{\ell}^{T} / \partial T & \partial R_{\ell}^{T} / \partial H \\
\partial R_{\ell}^{H} / \partial T & \partial R_{\ell}^{H} / \partial H
\end{array}\right)_{\substack{T=T^{*} \\
H=H^{*}}}\]

$M$ represents how small changes in T and H affect the transformations. The eigenvectors of M are special directions in which the transformations act simply by stretching or shrinking, and they are combinations of $\Delta$T and $\Delta$H. Often, M is diagonal and does not mix T and H, which simplifies things. For now, we assume this is the case.\\

We write the eigenvalues of M (these are numbers associated with each eigenvector that tell us how much stretching or shrinking occurs in that direction) as:

\[\begin{aligned}
\Lambda_{\ell}^{t} & =\ell^{y_{t}} ; \\
\Lambda_{\ell}^{h} & =\ell^{y_{h}},
\end{aligned}\]

the Renormalisation Group transformation becomes

\[\left(\begin{array}{c}
t^{\prime} \\
h^{\prime}
\end{array}\right)=\left(\begin{array}{cc}
\Lambda_{\ell}^{t} & 0 \\
0 & \Lambda_{\ell}^{h}
\end{array}\right)\left(\begin{array}{l}
t \\
h
\end{array}\right) .\]

Before proceeding further with the derivation of power-law scaling. We will define the correlation length $\xi$ in Renormalisation Group theory. This gives a sense of how far apart two points can be while still influencing each other.\\

Mathematically, the correlation function typically decays exponentially for distances larger than the correlation length:
\[C(r) \approx exp(\frac{-r}{\xi})\]

where C(r) is the correlation function and r is the distance between two points.\\

Suppose under Renormalisation Group transformations, we zoom out by a factor of $\ell$, the correlation length effectively gets smaller by the same factor ($\xi \rightarrow \frac{\xi}{\ell}$).\\

Hence if we apply the Renormalisation Group transformation n times, the correlation length transforms as:
\[\xi(t, h)=\ell^{n} \xi\left(\ell^{n y_{t}} t, \ell^{n y_{h}} h\right)\]

The singular part of the free energy density then transforms according to:
\[\begin{aligned}
f(t, h) & =\ell^{-d} f\left(t^{\prime}, h^{\prime}\right)=\ell^{-n d} f\left(t^{(n)}, h^{(n)}\right) \\
& =\ell^{-n d} f\left(\ell^{n y_{t}} t, \ell^{n y_{h}} h\right),
\end{aligned}\]

This equation looks like the result we expected from the Kadanoff block spin argument, \ref{eq:scaling_relation}. Next, if we choose $\ell^n = b \cdot t^{-\frac{1}{y_t}}$, we obtain:
\[f(t, h)=t^{d / y_{t}} b^{-d} f\left(b, h / t^{y_{h} / y_{t}}\right)\]

The scaling behaviour of the singular part of the free energy density is typically written as:
\[f_{s}(t, h)=|t|^{2-\alpha} F_{f}\left(h /|t|^{\Delta}\right)\]
with 
\[
F_{f}(x) \equiv f_{s}(1, x)
\]
This is a key result known as the static scaling hypothesis, which is a central assumption in the Renormalisation Group analysis of critical phenomena. Where $\alpha$ is the critical exponent associated with the heat capacity. More specifically, $2-\alpha$ is the scaling dimension of the free energy. $\Delta$ describes how the magnetic field $h$ scales with the reduced temperature $t$ in the critical region. \\

We then find:

\[\begin{aligned}
2-\alpha & =d \nu=\frac{d}{y_{t}} \\
\Delta & =y_{h} / y_{t} .
\end{aligned}\]

We now have a way to calculate the exponents $y_t$ and $y_h$, at least approximately, from the Renormalisation Group recursion relations.

\section[Understanding the Dynamics of Iterative Magnitude Pruning]{Understanding the Dynamics of Iterative Magnitude Pruning \cite[p.~4]{redman}}
The IMP process can be represented by the equation 
\[\mathcal{I} \mathcal{L}(\mathbf{a}, \boldsymbol{\theta})=\mathcal{L}\left(\mathbf{a}^{\prime}, \mathcal{T} \boldsymbol{\theta}\right)=\mathcal{L}\left(\mathbf{a}^{\prime}, \boldsymbol{\theta}^{\prime}\right)\]
Where the new set of parameters, $\boldsymbol{\theta}^{\prime}$, are given by an operator $\mathcal{T}$ and result in a new set of activations, $\mathbf{a}^{\prime}$.\\

In IMP, $\mathcal{T}$ is a combination of two operators: a masking operator $\mathcal{M}$ and a refining operator $\mathcal{F}$. This means that $\mathcal{T}=\mathcal{F} \circ \mathcal{M}: \mathbb{R}^{N} \rightarrow \mathbb{R}^{N}$, where $N$ is the number of parameters in the DNN. The pruning procedure used determines the definition of $\mathcal{M}$ (e.g. magnitude pruning), while the refinement procedure used defines $\mathcal{F}$. \\

For $n$ iterations of $\mathcal{I}$, the final DNN is given by 
\[\mathcal{I}^{n} \mathcal{L}(\mathbf{a}^{(0)}, \boldsymbol{\theta}^{(0)})=\mathcal{L}(\mathbf{a}^{(n-1)}, \mathcal{T}^{n} \boldsymbol{\theta}^{(0)})=\mathcal{L}(\mathbf{a}^{(n-1)}, \boldsymbol{\theta}^{(n-1)})\]
This creates a path in parameter space, $\boldsymbol{\theta}^{(0)} \rightarrow \boldsymbol{\theta}^{(1)} \rightarrow \ldots \rightarrow \boldsymbol{\theta}^{(n-1)}$, called the IMP flow, determined by the eigenvectors of $\mathcal{T}$ and their eigenvalues.

\subsection{Power-law scaling in IMP}
Rosenfeld et al. (2020) \cite{ROSENFELD} recently discovered a similarity between universality in renormalization group (RG) and Lottery Ticket Hypothesis theories when they analysed the pruning of a DNN using IMP. They found that when the density (the percentage of remaining parameters) of the DNN falls within a certain range, $d_{L}<d<d_{C}$, the error of the DNN follows a power-law relationship, which can be expressed as $e \sim\left(d_{C}-d\right)^{-\gamma}=\Delta d^{-\gamma}$, where  $\gamma$ is the critical exponent.

\subsection{IMP flow}
To understand how IMP flow works, we need to find the eigenfunctions of the aforementioned operator $\mathcal{T}$. By examining the eigenvalues associated with these eigenfunctions, we can determine which directions are relevant and irrelevant. Note this is analogous to the section on fixed point analysis in Renormalisation Group theory.\\

For spin systems, Renormalisation Group flow is studied in the space of coupling constants. Where the coupling constants are assumed the same for all spins, which greatly reduces the dimensionality of the parameter space. However NNs don't set the parameters of the same type to the same value, hence we need to study the IMP flow in directly by estimating the relative "influence" the parameters of a given layer have on the full NN. This can be done by considering the total remaining parameter magnitude that remains in layer $i$ after $n$ applications of IMP \cite[p.~6]{redman}:\\
\[
M_i(n)=\frac{\sum_{j=1}^{N^{(i)}}\left|m_j^{(i)}(n) \cdot \theta_j^{(i)}(n)\right|}{\sum_{k=1}^N\left|m_k(n) \cdot \theta_k(n)\right|} 
\]

Here $N^(i)$ is the number of parameters in layer $i$ and $m^{(i)} \in \{0, 1\}^{N^{(i)}}$ is the pruning mask. The dot product of the parameters with the pruning mask makes sure only the non-pruned weights are considered.\\

If we are to consider $M_i(n)$ as eigenfunctions of the IMP operator they should scale exponentially with respect to the number of IMP iterations. As $M_{i}(n+1)=\mathcal{T}M_{i}(n)=\lambda_{i}M_{i}(n)=\lambda_{i}^{n+1}M(0)$. We can drive $\lambda_{i}$ as:\\
\[\lambda_{i}=\frac{M_{i}(n+1)}{M_{i}(n)}\]

In Redman et al, they found that $M_{i}(n+1)$ is appropriate to be considered as an eigenfunction due to its well-defined nature and the standard error of the mean of $\lambda_{i}$ is less than 5\%. The degree of coarse-graining ($x \in (0,1)$) at each iteration of IMP affects the magnitude of the eigenvalues. Therefore we are interested in the quantity $\sigma$:
\begin{equation}
\lambda_{i} \sim c^{\sigma_{i}}
\label{eq:SIGMA}
\end{equation}

Where $\sigma$ is invariant to the choice of c and taking $log_{c}(\lambda_{i})$ gives $\sigma_{i}$. Here c is defined as $\frac{1}{1-x}$. We want to compare across models that prune using different c values, we will report $\sigma_{i}$. Using this we have:

\begin{enumerate}[label=\arabic*.]
  \item Relevant directions, which have $\lambda_i > 1$, have $\sigma_i > 0$.
  \item Irrelevant directions, which have $\lambda_i < 1$, have $\sigma_i < 0$.
\end{enumerate}

The Lottery Ticket Hypothesis suggests that DNNs can be reduced to a low-dimensional subspace of important parameters during training, and this subspace can be used to transfer winning tickets between different DNN models. The Renormalisation Group theory provides tools for finding these subspaces and comparing them between models, by analyzing the eigenvectors and eigenvalues of the transformation matrix. The models that have the same eigenvectors with eigenvalues greater than 1 are said to have the same relevant parameters, making it possible to know if winning tickets can be transferred between them without additional experiments. However, having distinct relevant directions does not mean that the tickets cannot be transferred, but it suggests that the models have different properties and will be differently affected by the RG.

\section{Connection between Renormalisation Group and IMP}
The IMP and Renormalisation Group operator are techniques for simplifying complex physical systems. IMP "sparsifies" neural networks, while Renormalisation Group carries out coarse-graining on spin systems. By reducing the number of degrees of freedom in the original system, the coarse-grained spin system becomes simpler and more manageable. Both techniques show power-law scaling and have unique flows that allow us to understand the relevant components of the system that determine certain macroscopic behaviour.\\

Table 1. Showing analogous quantities in Renormalisation Group and IMP theory \cite[p.~2]{redman}.
\[\begin{tabular}{cc}
\hline $\mathbf{R G}$ & IMP \\
\hline Spins $\left(s_{i}\right)$ & Unit activations $\left(a_{i}\right)$ \\
Coupling constants $\left(k_{i}\right)$ & Parameters $\left(\theta_{i}\right)$ \\
Hamiltonian $(\mathcal{H}[\mathbf{s}, \mathbf{k}])$ & Loss function $(\mathcal{L}[\mathbf{a}, \boldsymbol{\theta}])$ \\
\hline
\end{tabular}\] \\

Prior to the Redman et al 2021 \cite{redman}, there was no connection made between renormalisation group and IMP. However previous research had made a connection between renormalisation group theory and deep learning \cite{MEHTA} \cite{HENRY} \cite{CEDRIC}. One important such paper was Mehta et al \cite{MEHTA}, which is based on the idea of hierarchical organization, where understanding at one level is built upon the understanding at a lower level. In deep learning, each layer of the network captures features at a different level of abstraction, with input data at the bottom and the final output at the top. This is analogous to the renormalization group, where the behaviour of a system at larger scales (higher levels) is derived from the behaviour at smaller scales (lower levels).

For a different flavour, \

\subsection{Proof: IMP is a Renormalisation Group scheme}
In this section, we meticulously unpack and refine the proof sketch provided by Redman \cite[p.~5]{redman}, incorporating additional clarity and rectifying minor inaccuracies. Our primary aim is to demonstrate that IMP aligns with the conceptual underpinnings of a Renormalisation Group scheme. To this end, we scrutinize a single application of the IMP procedure as encapsulated by the $\mathcal{I}$ operator. Subsequently, we endeavour to validate that $\mathcal{I}$ fulfils all prerequisites necessary to qualify as a Renormalisation Group projection operator, thereby substantiating the claim that IMP can indeed be categorized as a Renormalisation Group operator.\\

We start by establishing a concrete correlation between IMP and the Renormalisation Group theory by showing that IMP satisfies the specifications of a Renormalisation Group scheme. For this purpose, we direct our attention to the projection operator, $\mathcal{P}$, tied with the Renormalisation Group operator. This projection operator, $\mathcal{P}$, facilitates the mapping of spins within a classical spin system, denoted as $s_{i}$, onto a coarser spin system, $s_{I}^{\prime}$. The operation is characterized by the following equation:

\[
\operatorname{Tr}_{\left\{s_{i}\right\}} \mathcal{P}\left(s_{i}, s_{I}^{\prime}\right) \exp \left[\mathcal{H}\left(s_{i}, \mathbf{k}\right)\right]=\exp \left[\mathcal{H}\left(s_{I}^{\prime}, \mathbf{k}^{\prime}\right)\right],
\]
Where $\operatorname{Tr}_{\left\{s_{i}\right\}}$ is the trace operator over the values that the $s_{i}$ can take (e.g. $\pm 1$ ).\\

The projection operator must have three properties:\\
1) $\mathcal{P}\left(s_{i}, s_{I}^{\prime}\right) \geq 0$  \\ 
2) $\mathcal{P}\left(s_{i}, s_{I}^{\prime}\right) \text { respects the symmetry of the system }$. \\ 
3) $\sum_{\left\{s_{I}^{\prime}\right\}} \mathcal{P}\left(s_{i}, s_{I}^{\prime}\right)=1 \text {. }$ \\

Our primary objective is to identify the projection operator linked to $\mathcal{I}$. To initiate this, we map the activations of all units, denoted as 'a', before and after the application of IMP. This approach is based on the similarity between the activations 'a' and the quantity $\mathbf{s}$. Concentrating on unit $j$ in layer $i$, we represent its activation with the following equation:

\[a_{j}^{(i)}=h\left[\sum_{k} g_{k}(\mathbf{a}, \boldsymbol{\theta})\right],\]

This equation signifies that the activation of unit $j$ in layer $i$, symbolized as $a_{j}^{(i)}$, is equivalent to the activation function $h$ applied to the sum of functions $g_{k}$. The functions $g_{k}$ delineate the influence of different parameters and activations of other units on $a_{j}^{(i)}$.\\

In a feedforward DNN, $g_{0}$ represents the impact of the bias of unit $j$ in layer $i$, which is given by $g_{0}=\theta_{j}^{(i)}$, and the weighted input from the previous layer is given by $g_{1}=\sum_{k=1}^{N^{(i-1)}} \theta_{j k}^{(i)} a_{k}^{(i-1)}$. Here $N^{(i-1)}$ is the number of units in layer $i-1$.\\

The IMP method modifies the parameters $\theta$ of a deep neural network using the operator $\mathcal{T}$, which is a combination of two other operators $\mathcal{M}$ and $\mathcal{F}$. The resulting activation of unit $j$ in layer $i$ after the application of $\mathcal{I}$ is given by:

\[a_{j}^{\prime(i)}=h\left[\sum_{k} g_{k}(\mathbf{a^{\prime}}, \mathcal{F} \circ \mathcal{M} \boldsymbol{\theta})\right],\]

The projection operator $\mathcal{P}$ associated with $\mathcal{I}$ is defined as:\\

\[\mathcal{P}\left(a_{j}^{(i)}, a_{j}^{\prime(i)}\right)=\prod_{j=1}^{N} \delta\left\{a_{j}^{\prime(i)}-h\left[\sum_{k} g_{k}(\mathbf{a^{\prime}}, \mathcal{F} \circ \mathcal{M} \boldsymbol{\theta})\right]\right\}\]

Remarkably, this projection operator fulfills all three properties essential for a Renormalisation Group projection operator:

1) The product of Kronecker delta functions within $\mathcal{P}$ guarantees its non-negativity. \\
2) The Renormalisation Group projection operator, in preserving the inherent symmetry of the system, avoids introducing any new terms or couplings that did not exist originally. In the context of IMP, the operator merely removes the connections between units instead of introducing new forms of interaction. This preservation ensures the system's behaviour remains unaffected until a layer collapse occurs. In a layer collapse, all the weights interconnecting two layers are nullified. Preserving the original system symmetry is vital as pruning should not alter the fundamental nature of the model under investigation. \\
3) In the context of the IMP method, this property can be satisfied by fixing the ordering of test and training samples during each epoch by setting a fixed random seed. This ensures that the masking and refining operations defined in the projection operator $\mathcal{P}\left(a_{j}^{(i)}, a_{j}^{\prime(i)}\right)$ are deterministic and produce unique results. Then $\sum_{\left\{s_{I}^{\prime}\right\}} \mathcal{P}\left(s_{i}, s_{I}^{\prime}\right)=1 \text {. }$.\\

These observations imply that the Renormalisation Group theory serves as an apt language to examine IMP.\\

\emph{Note:} Within the domain of artificial neural networks, the term "layer collapse" refers to a scenario where all the weights that connect two layers within the network become null, effectively disrupting the information passage through these layers. Consequently, these two layers virtually merge into one, reducing the network's total layers. When such an event occurs, the loss function and the activations of the units may undergo significant changes, potentially affecting the network's training and performance.\\

\section{Identifying Gaps in the Connection Between Renormalisation Group Theory and IMP}
While the connection between Renormalisation Group theory and Iterative Magnitude Pruning (IMP) has been established, several gaps remain that need further investigation. The following are a few key areas that require additional exploration:
\begin{itemize}
\item \textbf{Lack of Correlation Function Analogy:} There is currently no clear analogy for the correlation function as used in Renormalisation Group theory within the IMP framework. We speculated that this may be related to the degree of correlation among the activations of different units for a given data set.
\item \textbf{Uniform Coupling Constants versus Diverse Parameters:} Renormalisation Group theory assumes that coupling constants are identical for each spin. In contrast, most Deep Neural Networks (DNNs) don't assign all parameters of the same type to the same value. This discrepancy poses a potential issue for drawing direct parallels between the two theories.
\item \textbf{Absence of Effective IMP Theory:} At present, there is no comprehensive theory that outlines what percentage of IMP is optimal for a model to maintain its performance. Correspondingly, the Renormalisation Group theory doesn't offer a similar concept. The development of such a theoretical framework would provide greater insight into the application and limitations of IMP within DNNs.
\item \textbf{Adaptability to Various Neural Network Architectures:} While IMP has been applied to different types of DNNs, it would be beneficial to investigate how the connection to Renormalisation Group theory extends to other types of neural network architectures, such as recurrent neural networks or generative adversarial networks.
\item \textbf{Inclusion of Other Pruning Strategies:} IMP is just one of many pruning strategies for DNNs. It's worth investigating how Renormalisation Group theory could be applied to other pruning strategies and whether the current connection could be generalized.
\item \textbf{Role of Initialisation:} The initialisation of the neural network weights plays a crucial role in the Lottery Ticket Hypothesis and hence in IMP. However, how this aspect correlates with the Renormalisation Group theory is not entirely understood and warrants further study.
\end{itemize}

\section{Objective and Design of the Experiments}
We have chosen two Hamiltonian neural networks (HNNs)\cite{HNN} from "Hamiltonian Neural Networks for Solving Equations of Motion" by M. Mattheakis \cite{marios}. These networks have been designed to solve equations of motion for the Non-linear Oscillator system and a chaotic Hénon-Heiles dynamical system. Although both neural networks are composed of three layers, the last layer in the Hénon-Heiles system is twice the size of the final layer in the Non-linear Oscillator system.\\

Our experiment involves the application of the Iterative Magnitude-based Pruning (IMP) method to each layer of both systems, as well as the systems as a whole. Our primary focus is to investigate whether IMP manifests the power-law scaling predicted by renormalisation group theory and to scrutinise the associated critical exponents. To facilitate this, we will calculate the $\sigma$ (as defined in Eq. \ref{eq:SIGMA}) for each layer. This will enable us to probe the similarities between the two systems and their architectural design, allowing us to anticipate the potential for transferability of their "winning tickets". Our ultimate objective is to identify these winning tickets for the two systems and evaluate whether their interchangeability aligns with our expectations based on the critical exponents and sigma values.\\

We anticipate some degree of transferability between the Non-linear Oscillator system and the Hénon-Heiles system. This expectation stems from the fact that both systems are non-linear dynamical systems \cite{strogatz1994nonlinear} embodying energy conservation principles, making them ideal for modelling various physical phenomena. Such similarities suggest that these systems could belong to the same equivalence class under the purview of Renormalisation Group theory.\\

For those interested in a deeper exploration of our study, our codebase and raw data are available at \cite{CODE}. While the universality of winning tickets among neural networks has been the subject of past studies, our work is novel in its specific application to Hamiltonian neural networks employed to solve differential equations \cite{dog23_3,chen2018neural,dog23_2}.

\subsection{Introduction to Hamiltonian Neural Networks}
Hamiltonian neural networks (HNNs), as proposed in the literature \cite{marios,HNN,dog20,dog20_npbe, dogred20_2}, introduce a novel approach to solving differential equations that describe dynamical systems. The Hamiltonian is a function that encapsulates the total energy (kinetic plus potential) of a physical system. By using the Hamiltonian as a guiding principle, HNNs learn to predict the evolution of a system over time.\\

The HNN architecture is comprised of neural networks that approximate the kinetic and potential energies of the system, as well as the gradients of these energies with respect to position and momentum variables. It's structured to inherently conserve the Hamiltonian, which is a desirable property in many physics problems where energy conservation \cite{dog23} is paramount. This conservation property can lead to more accurate and stable simulations compared to traditional neural networks that do not conserve energy.

\subsection{IMP experiment 1.  Nonlinear Oscillator}

The experiment is with a one-dimensional nonlinear oscillator with hamiltonian:
\begin{align}
H(x, p) &= \frac{p^2}{2} + \frac{x^2}{2} + \frac{x^4}{2},
\end{align}

We assume that the natural frequency and the mass of the oscillator are considered to be unity. The corresponding equations that govern motion are:
\begin{align}
\dot{x} &= p, & \dot{p} &= -(x+x^3)
\end{align}

The loss function is the mean squared error:\\
\[L=\frac{1}{K} \sum_{n=1}^K\left[\left(\dot{\hat{x}}^{(n)}-\hat{p}^{(n)}\right)^2+\left(\dot{\hat{p}}^{(n)}+\hat{x}^{(n)}+\left(\hat{x}^{(n)}\right)^3\right)^2\right]\]

The hyper-parameters of the neural network when getting trained were set the same as in the paper. This neural network has 3 layers. The first and the hidden layer each have 50 neurons, and the output layer has 2 neurons. We use a learning rate of $8 \times 10^{-3}$ and trained for $5 \times 10^{4}$ epochs.\\

I ran IMP pruning experiments at $1\%$, $5\%$ and $10\%$, where we only pruned certain layers of the neural network and not the whole model. Pruning at different percentages is useful as it helps us investigate what percentage IMP is most effective when we carry out pruning of the entire model later. We start by pruning individual layers rather than the full model to learn about the layer-specific complexity as some layers might contain more redundancy or less critical information than others, which may warrant different pruning strategies.\\

The graphs below are average across many runs of the experiment until the graph stopped changing significantly.\\

\begin{figure}[H]
    \centering
    \begin{subfigure}[b]{0.45\textwidth}  
        \includegraphics[width=\textwidth]{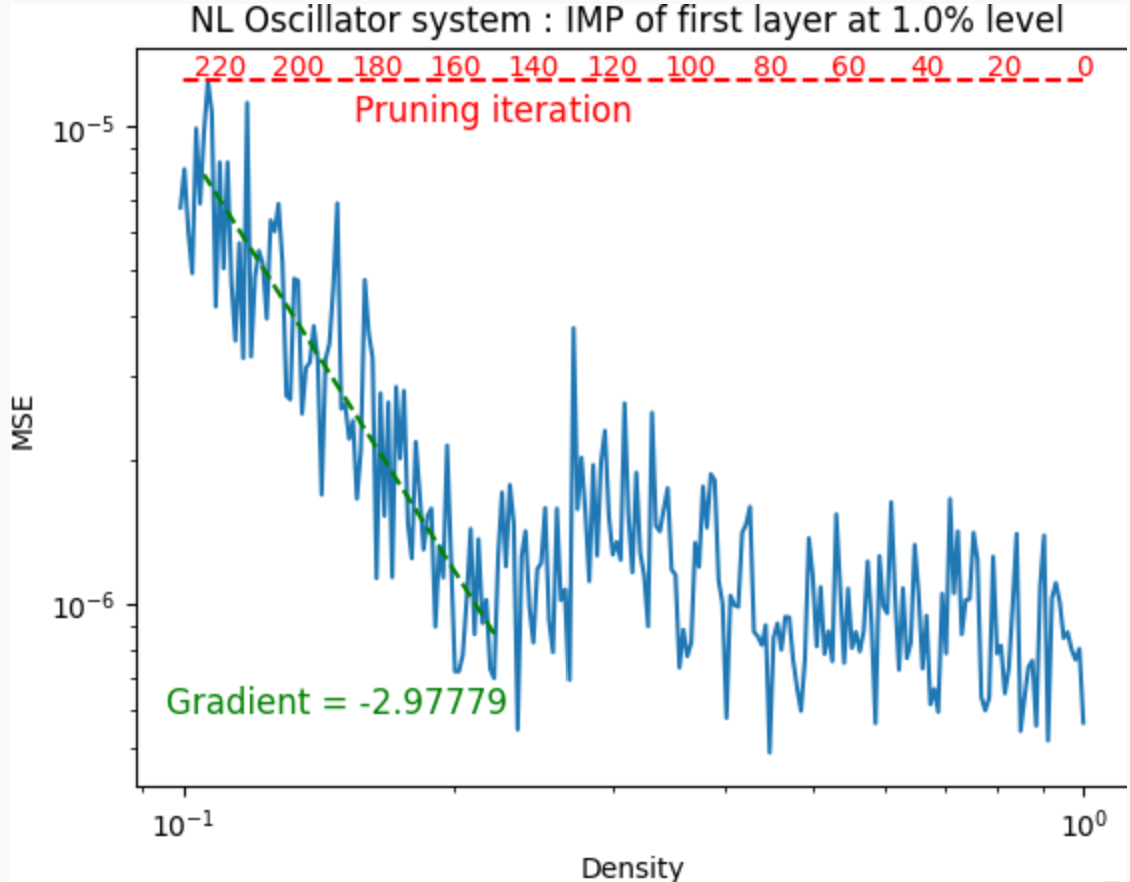}
        \caption{}  

    \end{subfigure}
    \hfill  
    \begin{subfigure}[b]{0.45\textwidth}  
        \includegraphics[width=\textwidth]{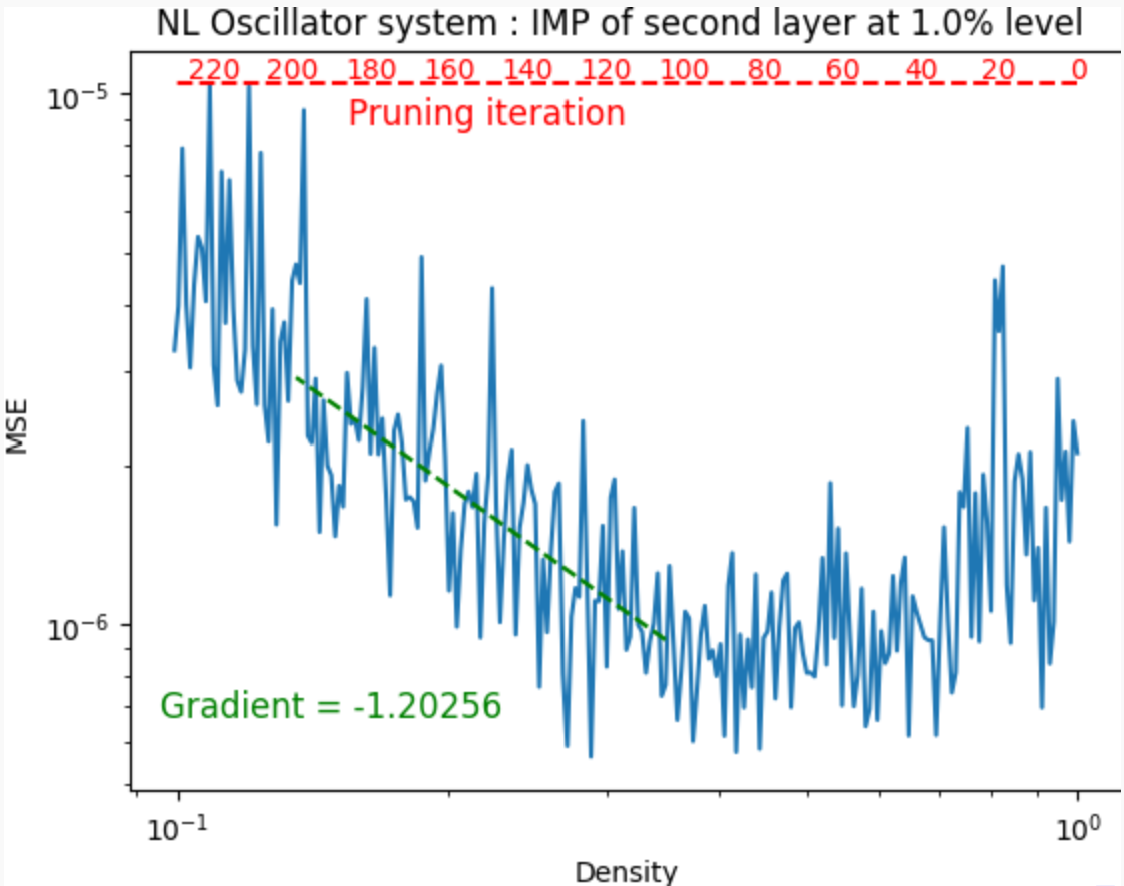}
        \caption{}  
    \end{subfigure}
    
    \begin{subfigure}[b]{\textwidth}  
        \centering  
        \includegraphics[width=0.45\textwidth]{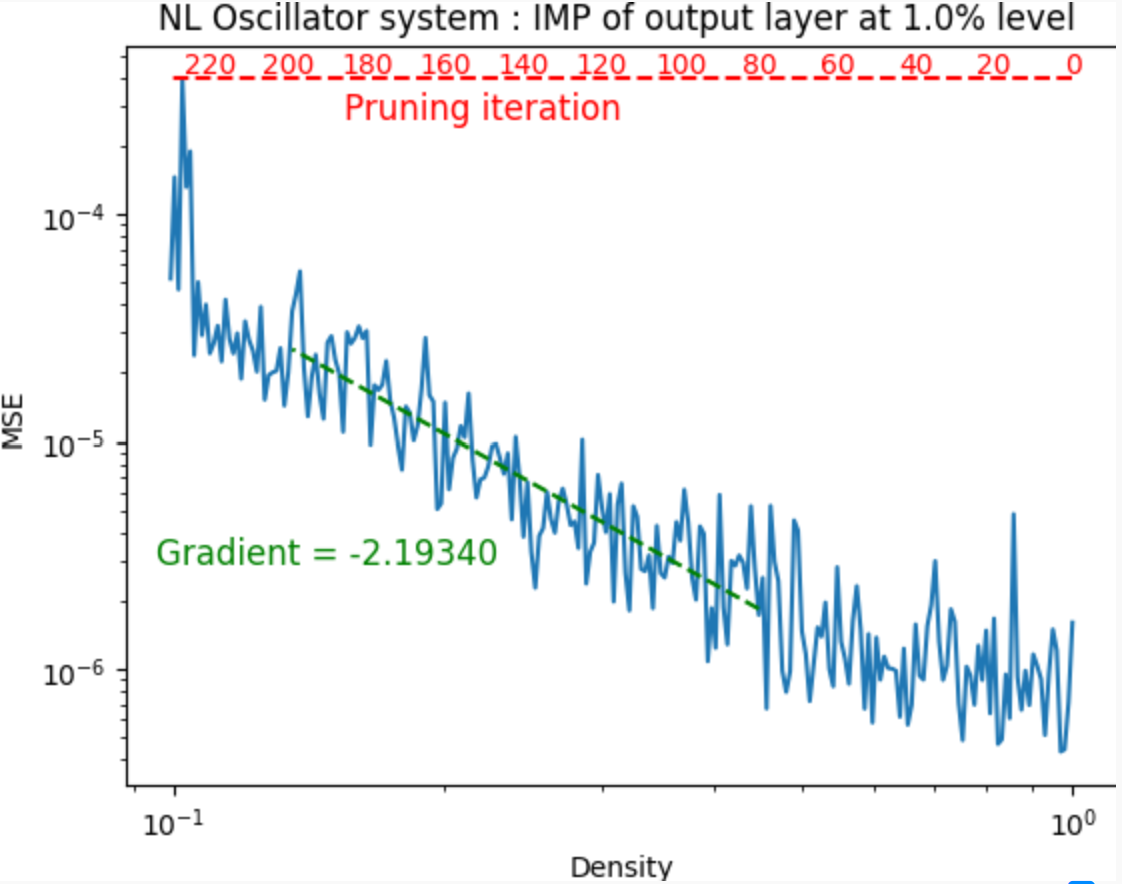}
        \caption{}  
    \end{subfigure}
    \caption{IMP at 1\% performed on each layer separately}
    \label{fig:1}
\end{figure}

From Figure \ref{fig:1} (a) and (b) we notice that we can prune 60\% of the input layer and hidden layer before we start to see the power-law scaling that we expect from IMP being a renormalisation group. The effectiveness of IMP is very apparent for the hidden layer as the MSE actually consistently decreases while we prune 30\% of the hidden layer. This layer-specific behavior aligns with the findings of Zhang et al. \cite{zhang2019alllayers}, who proposed a distinction between 'robust' and 'critical' layers in a deep neural network. Our first and hidden layers may be viewed as 'robust', with redundant representation capability, enabling the remaining neurons to compensate even after significant pruning. Conversely, the output layer, which begins to show power-law scaling after 25\% pruning, could be considered more 'critical' \cite{zhang2019alllayers}.

\begin{table}[H]
\centering
\begin{tabular}{|c|c|}
\hline
 & Critical exponent at 1\% IMP\\ 
\hline
NL Oscillator neural network only input layer pruning  & 2.9777 \\ 
\hline
NL Oscillator neural network only hidden layer pruning  & 1.2025 \\ 
\hline
NL Oscillator neural network only output layer pruning  & 2.1934 \\ 
\hline
\end{tabular}
\caption{Critical exponents for full model pruning. We work these out by taking the negative of the linear regression on the power-law scaling critical region as explained in section 4.1.}
\label{tab:4}
\end{table}

The critical exponent of the input layer and the output layer is significantly larger than that of the hidden layer. This suggests that the input layer is the most important to the model's predictions and the output layer is the second most important to the model predictions. This is because the input layer is the primary feature extractor from the input and the output layer is the decision-making layer. \\

Now the results from layer-specific pruning for IMP at 5\%.

\begin{figure}[H]
    \centering
    \begin{subfigure}[b]{0.45\textwidth}  
        \includegraphics[width=\textwidth]{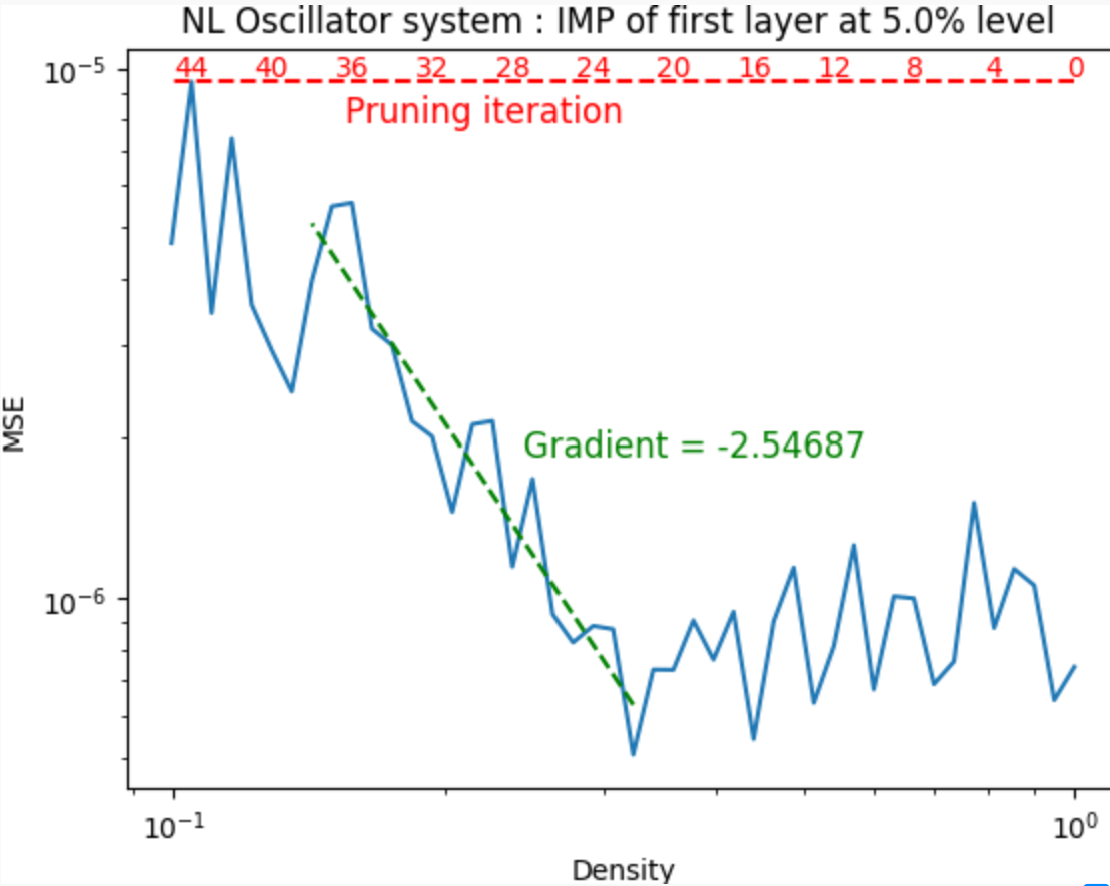}
        \caption{}  

    \end{subfigure}
    \hfill  
    \begin{subfigure}[b]{0.45\textwidth}  
        \includegraphics[width=\textwidth]{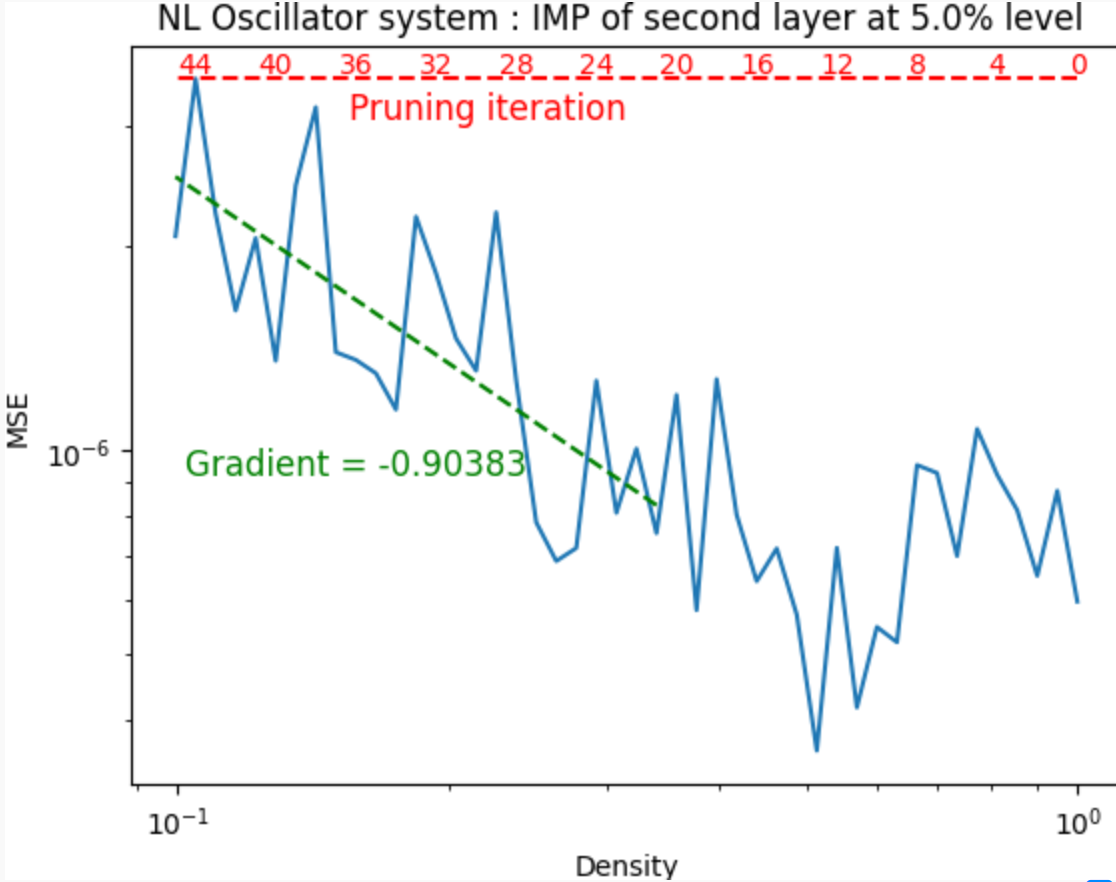}
        \caption{}  

    \end{subfigure}
    
    \begin{subfigure}[b]{\textwidth}  
        \centering  
        \includegraphics[width=0.45\textwidth]{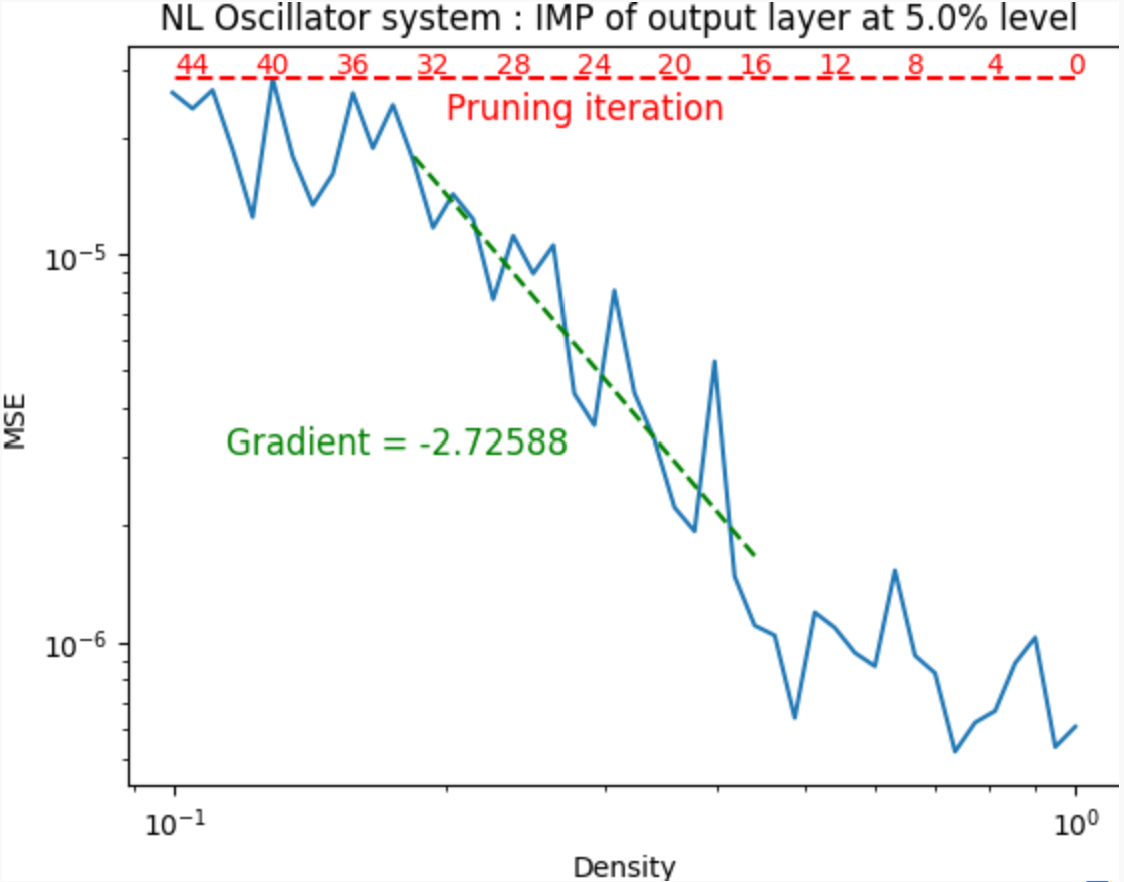}
        \caption{}  

    \end{subfigure}
    \caption{IMP at 5\% performed on each layer separately}
    \label{fig:2}
\end{figure}

The critical exponents have changed slightly as now the output layer has the largest critical exponent (2.7258) and the input layer has the second largest critical exponent (0.9038), This however could just be a bad estimation due to the small amount of data points. Overall, we still see a strong power-law scaling and the critical region of power-law scaling exists at similar densities as that of IMP at 1\%. \\

The results for IMP at 10\% level:

\begin{figure}[H]
    \centering
    \begin{subfigure}[b]{0.45\textwidth}  
        \includegraphics[width=\textwidth]{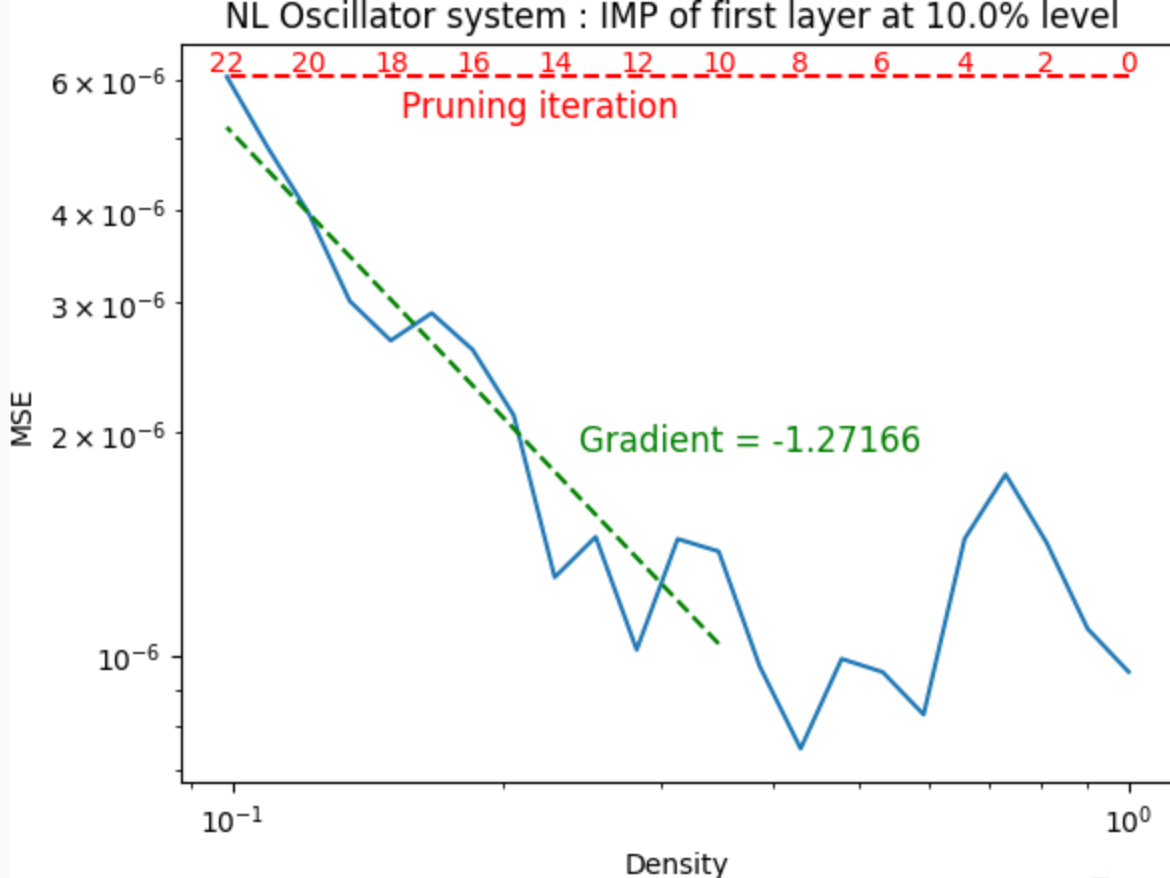}
        \caption{}  

    \end{subfigure}
    \hfill  
    \begin{subfigure}[b]{0.45\textwidth}  
        \includegraphics[width=\textwidth]{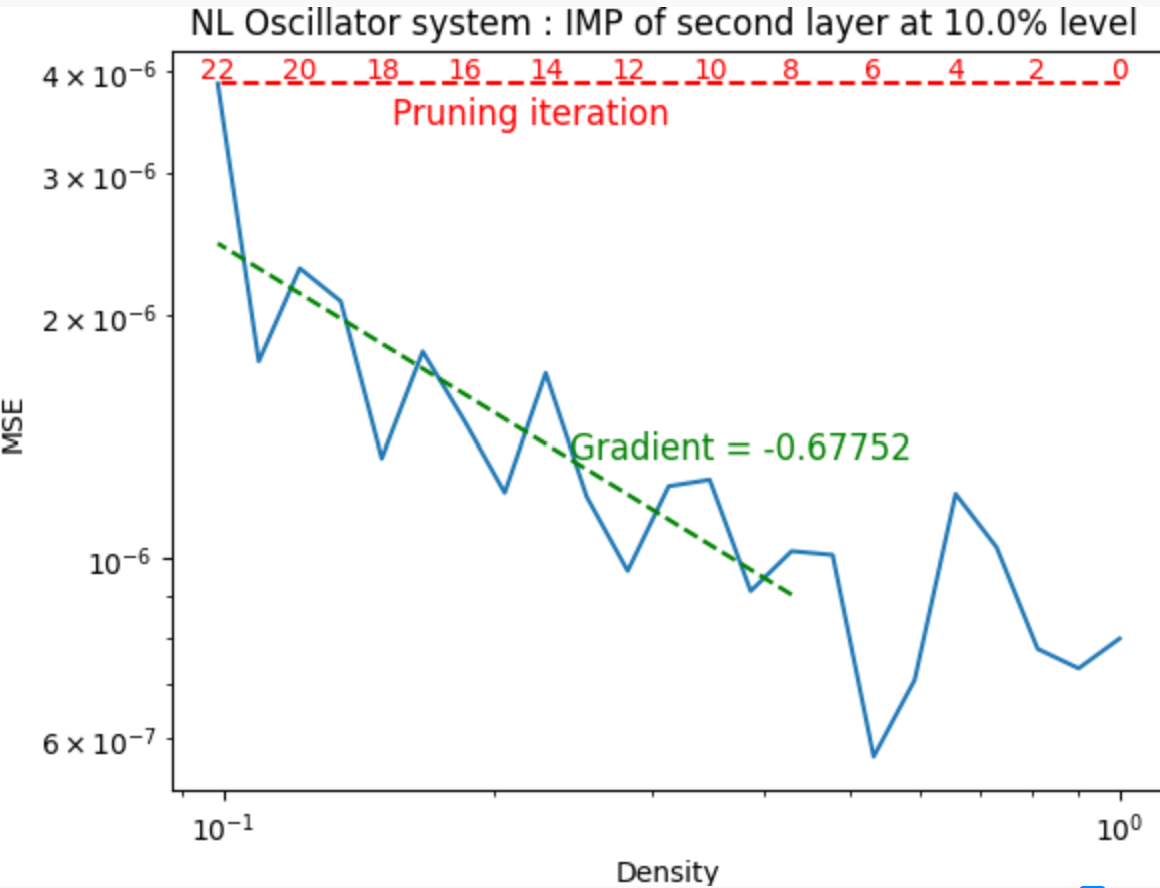}
        \caption{}  

    \end{subfigure}
    
    \begin{subfigure}[b]{\textwidth}  
        \centering  
        \includegraphics[width=0.45\textwidth]{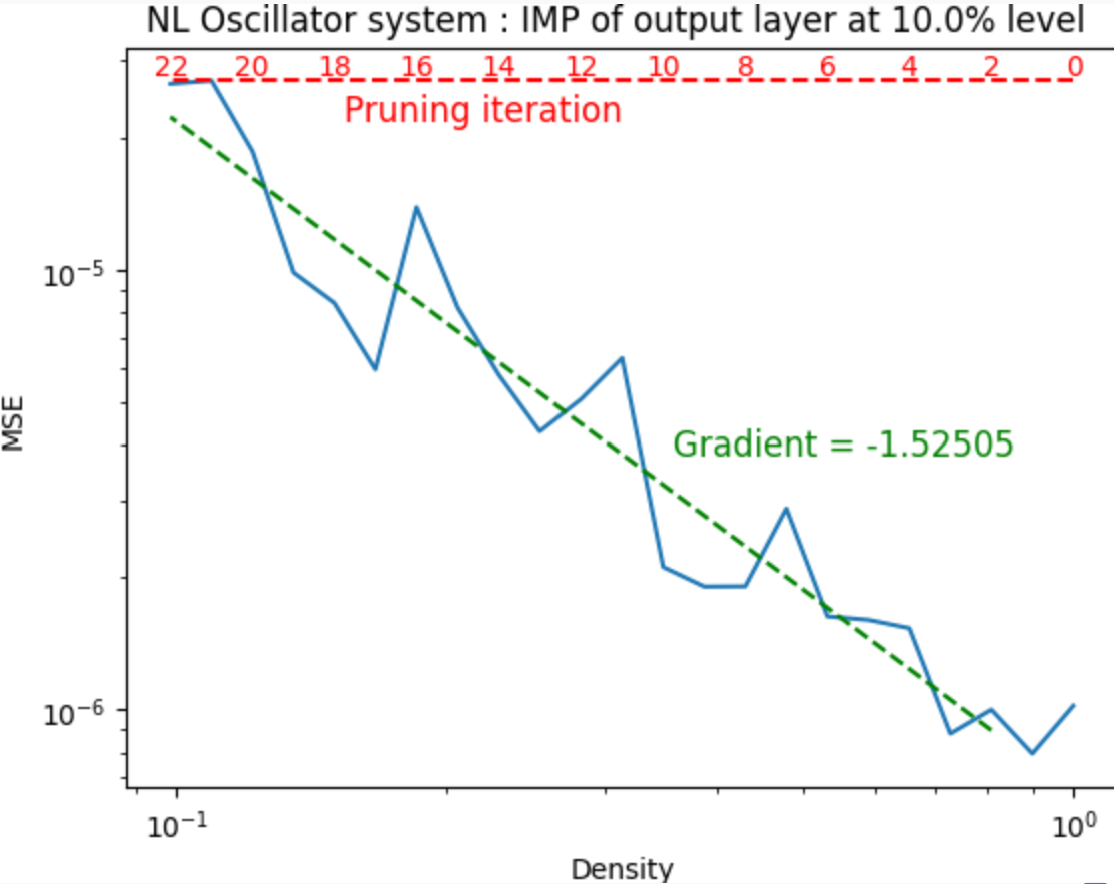}
        \caption{}  

    \end{subfigure}
    \caption{IMP at 10\% performed on each layer separately}
    \label{fig:3}
\end{figure}

Our results have provided empirical evidence that power-law scaling in IMP exists at 1\%, 5\% and 10\% levels. The critical exponents change with different levels of IMP but overall the input and output layer is the most important. IMP at 1\%, 5\% and 10\% took 230, 41 and 22 pruning iterations respectively to prune the layers by 90\%.\\

We notice that the critical densities where power-law scaling is observed remain similar across all IMP percentage levels. Furthermore IMP at 1\% level helps estimate the critical exponents most accurately due to the large number of data points available.\\

We decided that to investigate the transferability of winning tickets we will employ IMP at 1\% for experiments henceforth as even though this takes roughly 5 times the computational times as IMP at 5\% level, it can potentially lead to discovering more winning tickets due to greater pruning iterations.  \\

The pruning of the output layer demonstrated the most stability, evidenced by its mean square error (MSE) line adhering closely to the linear regression line across all pruning percentages. This intriguing finding implies that each layer of a deep neural network may exhibit different levels of tolerance to sparsity. Therefore, a more effective pruning strategy for the discovery of 'winning tickets' may involve pruning the smallest weights uniformly across all layers. This approach capitalizes on Iterative Magnitude Pruning (IMP) to maintain optimal model performance.\\

One must remember that in Deep Neural Networks (DNNs), layers are not standalone entities. They are intricately interconnected, with alterations to one potentially triggering ripple effects across others. In this regard, pruning the network as a whole, rather than selectively pruning individual layers, effectively accounts for these layer-to-layer interactions, thereby ensuring optimal performance.\\

When we apply the lens of renormalization group theory to the pruning process of a deep neural network, interesting parallels begin to emerge. If we consider the entirety of the network, this aligns with the 3D Ising model. The depth of the neural network can be likened to the layers of a 3D Ising model, almost as if they were sheets of 2D Ising models stacked upon each other. On the other hand, pruning specific layers in isolation resonates with the 2D Ising model. Given the complex interactivity among layers in a neural network, the 3D Ising model offers a more realistic analogy. Therefore, we posit that a more comprehensive pruning strategy encompassing the entire model would not only be more appropriate but also potentially more successful in uncovering those elusive 'winning tickets'.\\

Presented below is the average results graph for the pruning of the entire model:
\begin{figure}[H]
\centering
 \includegraphics[width=0.9\textwidth, height=0.35\textheight]{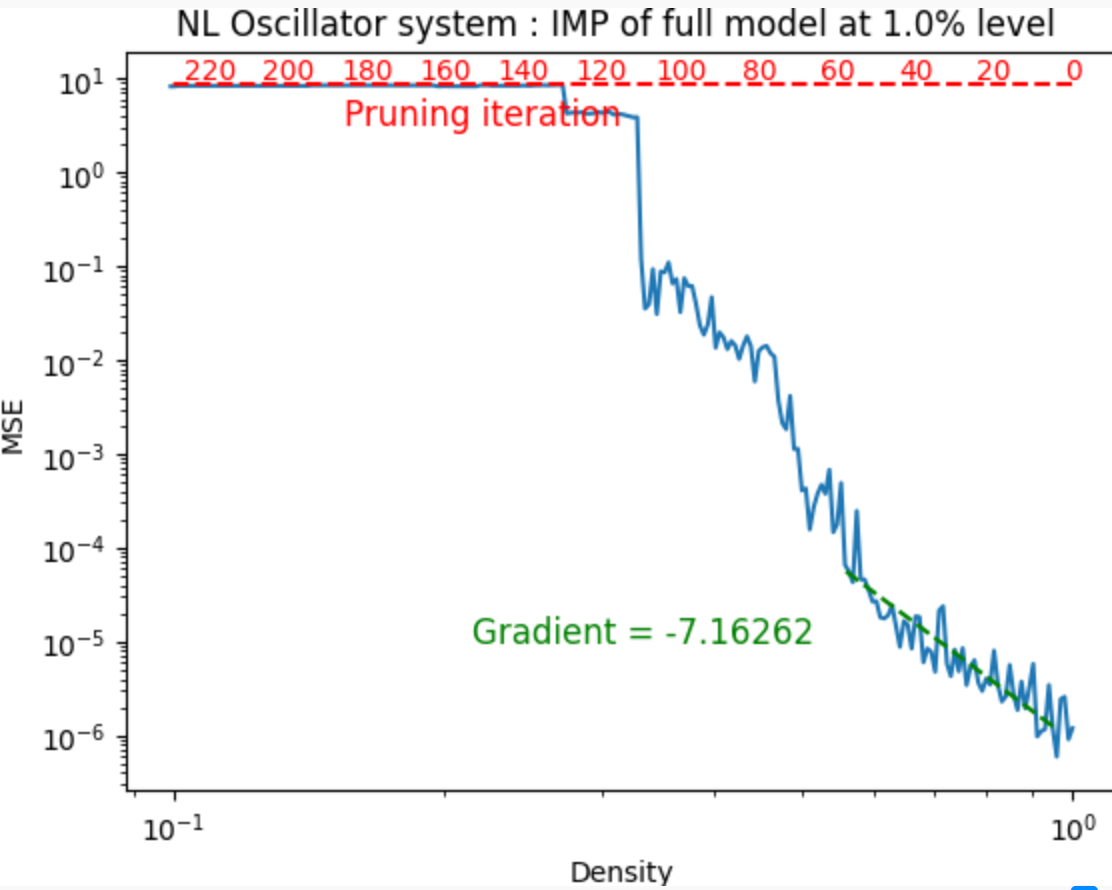} 
\caption{IMP the full NL oscillator model}
\label{fig:5}
\end{figure}

As observed in Figure \ref{fig:5}, we find that for densities below 0.9, the error begins to increase, exhibiting a power-law relationship. By scrutinizing each experiment's outcomes individually, we accumulate compelling empirical evidence to suggest that for densities below 0.9, winning tickets for the Nonlinear Oscillator model are non-existent, regardless of the initial parameters at which the IMP process is commenced.\\

This prompts us to shift our investigative lens from focusing exclusively on winning tickets to exploring the transferability of all subnetworks obtained from the original network. This includes subnetworks that may not necessarily show high performance on the original task but may exhibit interesting properties, such as improved performance upon transfer and fine-tuning on a different task as highlighted in the recent work by Fu et al. \cite{fu2023robust}. Thus, our investigation now expands to not just those tickets that win outright but all tickets with potential, including those that may only reveal their worth across different tasks.\\

Later in this discussion, we will derive the $\sigma$ for both the Nonlinear Oscillator and the Henon-Heiles (HH) system, as delineated in Section 7.2. This will enable us to ascertain the level of expected transferability between the winning tickets of both the Nonlinear Oscillator and HH system.

\subsection{IMP experiment 2. Chaotic Hénon-Heiles dynamical system}
Now we address the Hénon-Heiles system, which models a star's nonlinear motion around a galactic centre, confined to a plane. This system has four degrees of freedom in phase space, represented as z = (x, y, $p_x$, $p_y$). The system's Hamiltonian is:
\begin{align}
H(x, y, p_x, p_y) &= \frac{1}{2}(p_{x}^{2}+p_{y}^{2})+\frac{1}{2}(x^{2}+y^{2})+(x^{2}y-\frac{y^{3}}{3}),
\end{align}

Hamilton's equations lead to a system of nonlinear differential equations:
\begin{align}
\dot{x} &= p_{x} & \dot{y} &= p_{y} \\
\dot{p_{x}} &= -(x+2xy) & \dot{p_{y}} &= -(y+x^{2}-y^{2})
\end{align}

The loss function of the HH system's neural network is:
\[\begin{aligned} L & =\frac{1}{K} \sum_{n=0}^K\left[\left(\dot{\hat{x}}^{(n)}-\hat{p}_x^{(n)}\right)^2+\left(\dot{\hat{y}}^{(n)}-\hat{p}_y^{(n)}\right)^2\right. +\left(\dot{\hat{p}}_x^{(n)}+\hat{x}^{(n)}+2 \hat{x}^{(n)} \hat{y}^{(n)}\right)^2 \\ & \left.+\left(\dot{\hat{p}}_y^{(n)}+\hat{y}^{(n)}+\left(\hat{x}^{(n)}\right)^2-\left(\hat{y}^{(n)}\right)^2\right)^2\right] \end{aligned}\]

The hyper-parameters of the neural network again when getting trained were set the same as in the paper. This neural network has 3 layers. The first and the hidden layers have 50 neurons, and the output layer has 4 neurons. We use a learning rate of $8\times10^{-3}$ and trained for $2\times10^4$ epochs.

IMP of the HH system at 1\% gives:
\begin{figure}[H]
\centering
 \includegraphics[width=0.9\textwidth, height=0.35\textheight]{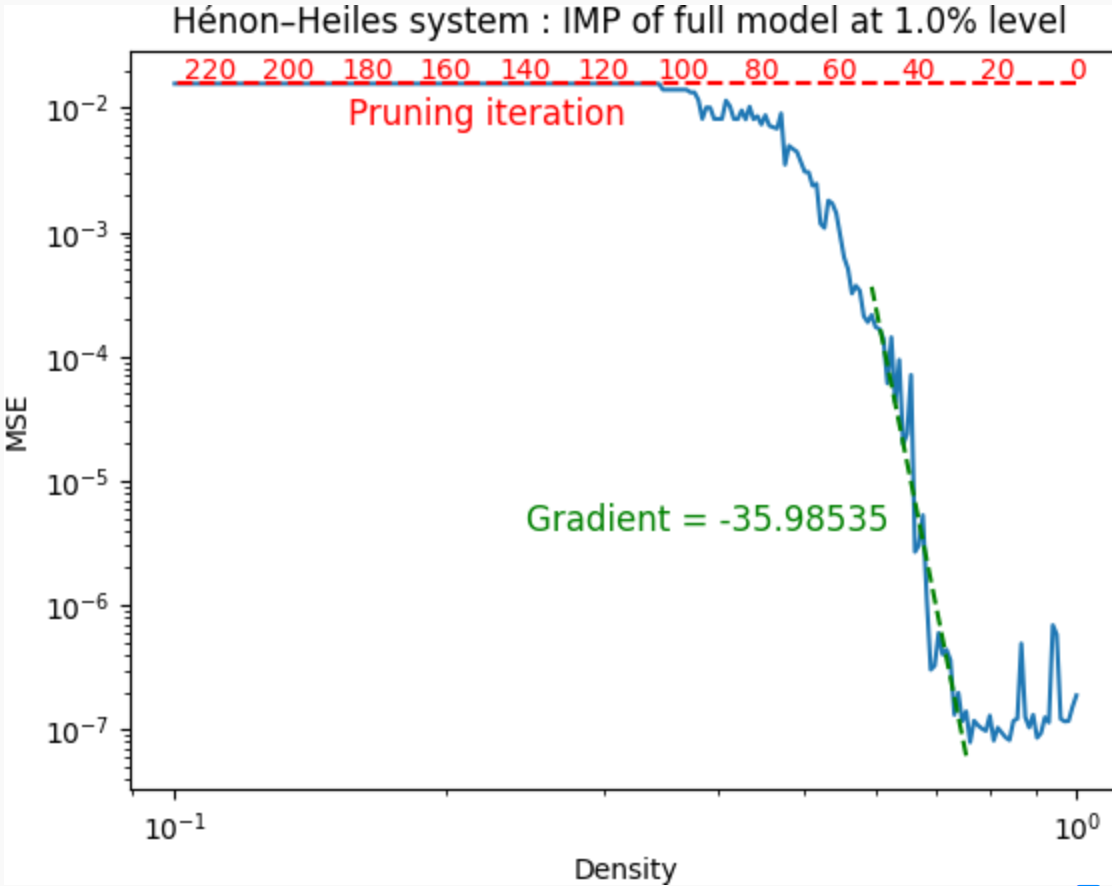} 
\caption{IMP the full HH system model}
\label{fig:6}
\end{figure}

Figure \ref{fig:6} shows that the HH system is able to produce winning tickets for densities between 1 and 0.75 whereas Figure \ref{fig:5} shows that NL oscillator system is only capable of producing winning tickets for densities 1 to 0.9. The power law scaling is also much steeper for HH system when compared to the NL oscillator (-35.9853 vs -7.1626). The distinctions between the two systems and architectures are beginning to manifest themselves within our graphical representations. Since now we are pruning the entire model, we can determine whether tickets are transferrable between them by comparing the $sigma$s of the two models as described in section 7.2.

\section{Tranferability of winning ticket between Hénon-Heiles system HNN and Nonlinear Oscillator HNN}
In this section, we will be conducting a pair of experiments to investigate the transferability of winning tickets between the Hénon-Heiles (HH) and Nonlinear Oscillator systems. Firstly, we will transfer masks derived from the Iterative Magnitude Pruning (IMP) process of the Nonlinear Oscillator to the HH system, evaluating their performance. We will then reciprocate this procedure by transferring the mask derived from IMP of the NL oscillator HNN to the HH system HNN.\\

To anticipate the outcomes of the transferability experiments, we will initially calculate the $\sigma$ values, as detailed in Section 7.2.

\begin{table}[H]
\centering
\begin{tabular}{|c|c|c|c|}
\hline
 & $\sigma_{1}$ & $\sigma_{2}$ & $\sigma_{3}$ \\ 
\hline
NL Oscillator neural network full model pruning  & 2.4778 & -0.1710 & 5.2449 \\ 
\hline
HH system neural network full model pruning  & 2.6447 & -0.1205 & 4.4276\\ 
\hline
\end{tabular}
\caption{Sigmas: approximating the influence of the parameters of each layer}
\label{tab:7}
\end{table}

Observing the table, it is evident that both $\sigma_{1}$ and $\sigma_{3}$ are positive for both models, thus corresponding to relevant directions. Conversely, $\sigma_{2}$ for both models is negative, indicating an irrelevant direction. From the perspective of Renormalisation Group flow, both models share similar relevant and irrelevant directions. \\

The magnitudes of $\sigma_{1}$ and $\sigma_{2}$ for both models exhibit concordance up to one significant figure. However, there is a difference in $\sigma_{3}$ for the Nonlinear Oscillator and the HH system. This discrepancy suggests potential variations in the scaling behaviour along this direction.\\

Given the similar relevant and irrelevant directions, we anticipate some degree of transferability between the two models. However, the variations in $\sigma_{3}$ could introduce unique behaviour during the transferability experiments. \\

\subsection{Transfer from HH system to NL Oscillator system in HNNs}
Transferring masks between the HH system and the NL oscillator faces a challenge due to the mismatch between the output layers of the two systems, with the NL oscillator having 2 neurons. To address this, we employ scaling, making transferability feasible by duplicating the mask of the NL oscillator when transferring the tickets to the HH system.\\

The training epochs differ between the NL oscillator neural network ($5 \times 10^{4}$) and the HH system neural networks ($2 \times 10^{4}$). This divergence arises from training each system to their respective natural convergence levels, rather than aligning them to a uniform convergence level. The HH system neural network naturally achieves a lower MSE (reaching $10^{-7}$) than the NL oscillator neural network (reaching $10^{-6}$). However, the transferability of the winning ticket is more linked to the network architecture \cite{LTH} — specifically, the initial structure and weights — rather than the nuances of the training process, such as learning rate and number of epochs. Consequently, differences in the training process do not adversely affect our research.\\

\begin{figure}[H]
\centering
\includegraphics[width=0.9\textwidth, height=0.35\textheight]{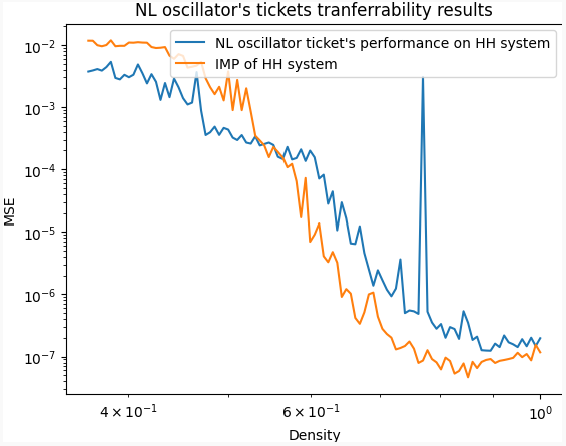}
\caption{Transferability of winning tickets: The plot showcases the comparative performance of winning ticket transfer from the Nonlinear Oscillator (NL) model to the Henon-Heiles (HH) system (blue line), against the standard performance of the iterative magnitude pruning (IMP) directly applied on the HH system (orange line).}
\label{fig:7}
\end{figure}

Initially, we notice a slight divergence between the two lines within a density range of 1 to 0.75. This disparity gradually intensifies and remains relatively constant throughout the majority of pruning iterations. Intriguingly, around a density of 0.65, the lines exhibit a brief period of intersection. Subsequently, the lines diverge once again, reverting to a similar spacing as observed earlier. In this latter stage, the transferred tickets showcase superior performance compared to the direct application of IMP on the HH system.\\

For densities between 1 and 0.75, the transfer of winning tickets proves successful. This outcome is noteworthy since winning tickets have traditionally been identified for high-dimensional problems with millions of parameters \cite{redman} \cite{LTH}. In contrast, we have demonstrated that winning tickets can be transferred for low-dimensional problems, like our two nonlinear dynamical problem Hamiltonian Neural Networks (HNNs), which contain fewer than 3000 parameters.\\

Interesting phenomena occur for densities below 0.7. For these densities, the MSE is at least tenfold that of the full HH system, rendering the tickets below a density of 0.7 ineligible as winning tickets. Interestingly, for densities lower than 0.65, the transferred tickets perform better on the HH system than direct IMP on the HH system.

\subsection{Transfer from NL Oscillator System to HH System in HNNs}

Addressing the discrepancy in the output layer of the Hénon-Heiles (HH) system, with 4 neurons, and the Nonlinear Oscillator (NL) system, with 2 neurons, is critical for successful mask transfer. To fit a 4 by 50 weights mask onto a 2 by 50 weight space, we truncate the second and fourth rows before initiating the transfer process.\\

\begin{figure}[H]
\centering
\includegraphics[width=0.9\textwidth, height=0.35\textheight]{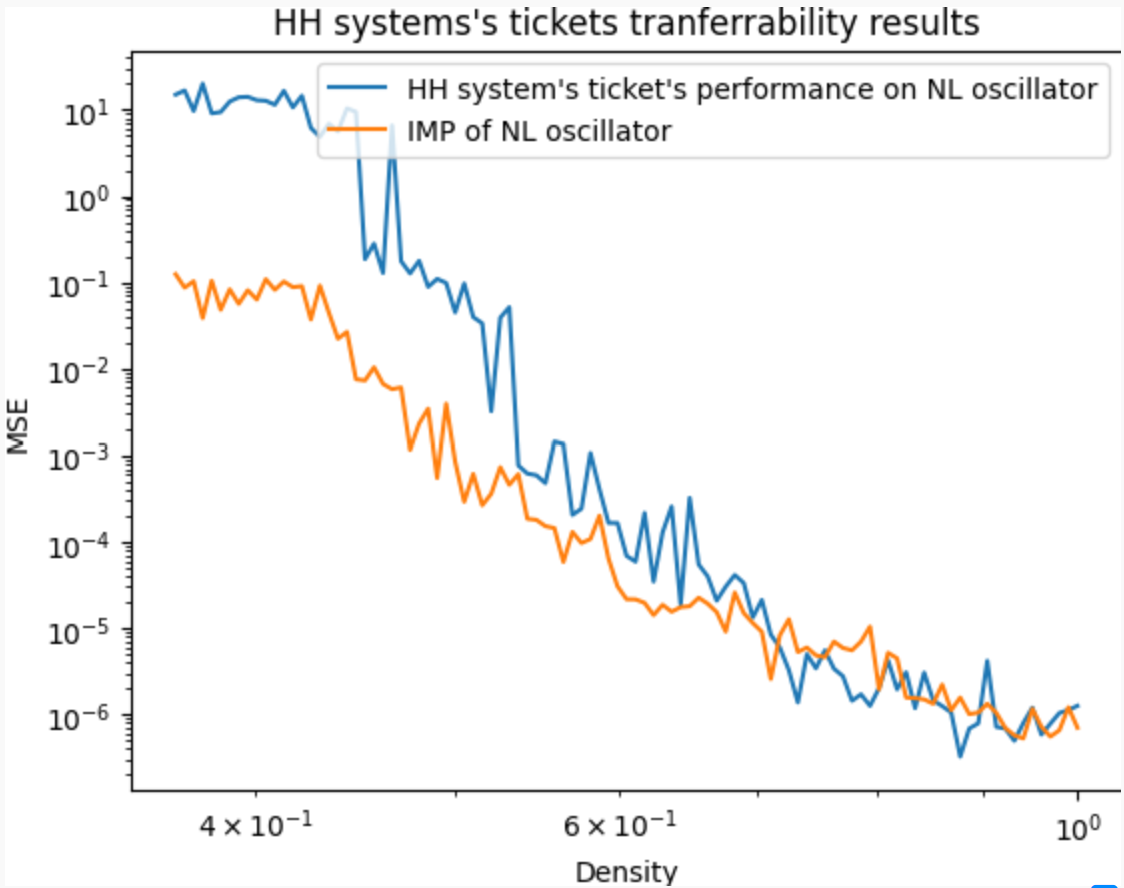}
\caption{Transferability of winning tickets from HH system to NL system: Performance comparison between direct iterative magnitude pruning (IMP) on the NL system and transferred tickets from HH system.}
\label{fig:8}
\end{figure}

We present our results on the transferability of masks from the HH system to the NL system in Figure \ref{fig:8}.

Interestingly, the transfer of tickets from the HH system to the NL oscillator displays superior efficacy compared to the reciprocal experiment, as indicated by the closer proximity of the performance curves.\\

Winning tickets are identified for densities ranging from 1 to 0.7. Even for densities below this range, the transferred tickets perform comparably to typical IMP performance on the NL oscillator, down to densities of 0.65. For densities lower than 0.65, the transferred tickets yield an MSE 10 to 20 times larger than the typical IMP of the NL oscillator.\\

The improved transferability of tickets from the HH system to the NL oscillator could potentially be attributed to the truncation of the output layer mask, which eliminates redundancy. This suggests that the transfer process may favour an absence of redundant information over the repetition of masks.

\section{Conclusion}

In this thesis, we embarked on a journey through the realms of the lottery ticket hypothesis (LTH), iterative magnitude pruning (IMP), and Hamiltonian Neural Networks (HNNs), within the context of low-dimensional physics problems. \\

Our work provides solid evidence supporting the existence of winning tickets in smaller neural networks used for low-dimensional problems. This observation, which complements and extends the findings of Frankle and Carbin's seminal work \cite{LTH}, broadens the scope of LTH. The research results also imply that the concept of winning tickets transcends the high-dimensional landscape where it was initially identified, permeating into the realm of lower-dimensional physics problems.\\

The application of iterative magnitude pruning to two distinctly different systems — the nonlinear oscillator and the Hénon-Heiles system — yielded fascinating insights into the relationship between the initial network architecture and the system it models. We observed that smaller is indeed often better when pruning, suggesting potential avenues for efficient model development and refinement.\\

Notably, we ventured into the largely unexplored territory of transferring winning tickets across different systems and model architectures. Our findings underscore the potential of such transfers and point to a rich seam of research possibilities.\\

Furthermore, our work lends new perspectives to understanding the LTH through the lens of renormalisation group theory, suggesting an intriguing connection between these two areas. The parallel drawn between the behaviour of winning tickets and the universality phenomena of renormalisation group theory underscores the underlying order beneath the seeming chaos of deep learning model training.\\

In conclusion, we hope our exploration inspires further research into the lottery ticket hypothesis, its manifestations in different domains, and its intriguing ties with renormalisation group theory. As we continue to unravel the mysteries of neural networks, we move closer to the creation of more efficient, effective, and transformative artificial intelligence technologies.

\section{Directions for Future Research}
In light of the findings and conclusions reached within this study, we propose the following promising directions for future research:

\begin{itemize}
\item \textbf{Optimizing Pruning Percentage:} While our study relied on a predetermined pruning rate, future research could explore methodologies for determining the optimal pruning percentage for a given model. The question arises whether smaller pruning percentages always yield better performance. As such, a systematic investigation into the effects of various pruning rates on the lottery ticket hypothesis and the transferability of tickets would be valuable.

\item \textbf{Adaptability of Masks Across Architectures:} We managed to transfer masks between models of differing architectures by duplicating the mask of the smaller architecture. However, the question remains whether other strategies may yield superior results. For instance, when transferring from a larger to a smaller network, would it be advantageous to truncate the mask, average it, or find a maximum? Alternatively, could we develop a mapping function to make this transfer more effective? And how should we handle the reverse scenario, transferring from a smaller to a larger system?

\item \textbf{Bridging the Gap between IMP and RG Theory:} As we mentioned in Section 7, our current work has only just begun to draw parallels between the lottery ticket hypothesis and the Renormalization Group (RG) theory. Additional research is required to further elucidate the connections and commonalities between these two paradigms.

\item \textbf{Restricting Pruning to Hidden Layers:} Our pruning strategy included the input and output layers. A worthwhile avenue for exploration is whether constraining pruning to just the hidden layers might yield superior performance in the identification of winning tickets.

\item \textbf{Exploring Different Architectures:} Our work to date has primarily focused on Hamiltonian Neural Networks (HNNs). Future work could expand this focus to include other architectures. In particular, the Deep Operator Network (DeepONet) architecture, introduced by Lu et al. \cite{Lu2021}, appears particularly promising. With its capacity for learning both explicit and implicit operators, DeepONet may offer substantial improvements in the efficiency and accuracy of solutions to the equations of motion in our HNN models.
\end{itemize}

We study the IMP flow in directly by estimating the relative "influence" the parameters of a given layer have on the full NN. This can be done by considering the total remaining parameter magnitude that remains in layer $i$ after $n$ applications of IMP \cite[p.~6]{redman}:\\
\[
M_i(n)=\frac{\sum_{j=1}^{N^{(i)}}\left|m_j^{(i)}(n) \cdot \theta_j^{(i)}(n)\right|}{\sum_{k=1}^N\left|m_k(n) \cdot \theta_k(n)\right|} 
\]

Here $N^(i)$ is the number of parameters in layer $i$ and $m^{(i)} \in \{0, 1\}^{N^{(i)}}$ is the pruning mask.

If we are to consider $M_i(n)$ as eigenfunctions of the IMP operator they should scale exponentially with respect to the number of IMP iterations. As $M_{i}(n+1)=\mathcal{T}M_{i}(n)=\lambda_{i}M_{i}(n)=\lambda_{i}^{n+1}M(0)$. We can drive $\lambda_{i}$ as:\\
\[\lambda_{i}=\frac{M_{i}(n+1)}{M_{i}(n)}\]

The degree of coarse-graining ($x \in (0,1)$) at each iteration of IMP affects the magnitude of the eigenvalues. Therefore we are interested in the quantity $\sigma$:
\begin{equation}
\lambda_{i} \sim c^{\sigma_{i}}
\label{eq:SIGMA}
\end{equation}

Where $\sigma$ is invariant to the choice of c and taking $log_{c}(\lambda_{i})$ gives $\sigma_{i}$.

\bibliographystyle{abbrvurl}
\bibliography{Abu}

\begin{thebibliography}{10}

\bibitem{dog23}
D.~Ba, A.~S. Dogra, R.~Gambhir, A.~Tasissa, and J.~Thaler.
\newblock Shaper: Can you hear the shape of a jet?
\newblock {\em Journal of High Energy Physics}, 2023(195), 2023.
\newblock URL: \url{https://doi.org/10.1007/JHEP06(2023)195}.

\bibitem{balwani2022zeroth}
A.~Balwani and J.~Krzyston.
\newblock Zeroth-order topological insights into iterative magnitude pruning,
  2022.
\newblock \href {http://arxiv.org/abs/2206.06563} {\path{arXiv:2206.06563}}.

\bibitem{CEDRIC}
C.~Bény.
\newblock Deep learning and the renormalization group, 2013.
\newblock \href {http://arxiv.org/abs/1301.3124} {\path{arXiv:1301.3124}}.

\bibitem{chen2018neural}
R.~T.~Q. Chen, Y.~Rubanova, J.~Bettencourt, and D.~K. Duvenaud.
\newblock Neural ordinary differential equations.
\newblock {\em Advances in Neural Information Processing Systems}, 31, 2018.

\bibitem{ELTH}
X.~Chen, Y.~Cheng, S.~Wang, Z.~Gan, J.~Liu, and Z.~Wang.
\newblock The elastic lottery ticket hypothesis.
\newblock {\em CoRR}, abs/2103.16547, 2021.
\newblock URL: \url{https://arxiv.org/abs/2103.16547}, \href
  {http://arxiv.org/abs/2103.16547} {\path{arXiv:2103.16547}}.

\bibitem{ARCHITECTURE}
E.~J. Crowley, J.~Turner, A.~J. Storkey, and M.~F.~P. O’Boyle.
\newblock Pruning neural networks: is it time to nip it in the bud?
\newblock {\em ArXiv}, abs/1810.04622, 2018.

\bibitem{dog20_npbe}
A.~S. Dogra.
\newblock A blueprint for building efficient neural network differential
  equation solvers.
\newblock {\em arXiv:2007.04433}, 2020.

\bibitem{dog20}
A.~S. Dogra.
\newblock Dynamical systems and neural networks, 2020.
\newblock \href {http://arxiv.org/abs/2004.11826} {\path{arXiv:2004.11826}}.

\bibitem{dog23_2}
A.~S. Dogra.
\newblock v tangent kernels.
\newblock {\em IAIFI Summer Workshop}, 2023.

\bibitem{dog23_3}
A.~S. Dogra, J.~B. Lai, and M.~Peev.
\newblock Neural network differential equation solvers allow unsupervised error
  analysis and correction.
\newblock 2023.

\bibitem{dogred20_2}
A.~S. Dogra and W.~T. Redman.
\newblock Local error quantification for efficient neural network dynamical
  system solvers.
\newblock {\em arXiv:2008.12190}, 2020.

\bibitem{dogred20}
A.~S. Dogra and W.~T. Redman.
\newblock Optimizing neural networks via koopman operator theory.
\newblock {\em Advances in Neural Information Processing Systems (NeurIPS)},
  33, 2020.

\bibitem{IMP}
B.~Elesedy, V.~Kanade, and Y.~W. Teh.
\newblock Lottery tickets in linear models: An analysis of iterative magnitude
  pruning.
\newblock {\em CoRR}, abs/2007.08243, 2020.
\newblock URL: \url{https://arxiv.org/abs/2007.08243}, \href
  {http://arxiv.org/abs/2007.08243} {\path{arXiv:2007.08243}}.

\bibitem{LTH}
J.~Frankle and M.~Carbin.
\newblock The lottery ticket hypothesis: Training pruned neural networks.
\newblock {\em CoRR}, abs/1803.03635, 2018.
\newblock URL: \url{http://arxiv.org/abs/1803.03635}, \href
  {http://arxiv.org/abs/1803.03635} {\path{arXiv:1803.03635}}.

\bibitem{morcos2019lotteryscale}
J.~Frankle, G.~K. Dziugaite, D.~M. Roy, and M.~Carbin.
\newblock The lottery ticket hypothesis at scale.
\newblock {\em CoRR}, abs/1903.01611, 2019.
\newblock URL: \url{http://arxiv.org/abs/1903.01611}, \href
  {http://arxiv.org/abs/1903.01611} {\path{arXiv:1903.01611}}.

\bibitem{fu2023robust}
Y.~Fu, Y.~Yuan, S.~Wu, J.~Yuan, and Y.~Lin.
\newblock Robust tickets can transfer better: Drawing more transferable
  subnetworks in transfer learning, 2023.
\newblock \href {http://arxiv.org/abs/2304.11834} {\path{arXiv:2304.11834}}.

\bibitem{GOLDENFELD}
N.~Goldenfeld.
\newblock {\em Lectures on Phase Transitions and the Renormalization Group}.
\newblock CRC Press, 2018.

\bibitem{DNN}
I.~Goodfellow, Y.~Bengio, and A.~Courville.
\newblock {\em Deep Learning}.
\newblock MIT Press, 2016.
\newblock \url{http://www.deeplearningbook.org}.

\bibitem{HNN}
S.~Greydanus, M.~Dzamba, and J.~Yosinski.
\newblock Hamiltonian neural networks.
\newblock {\em CoRR}, abs/1906.01563, 2019.
\newblock URL: \url{http://arxiv.org/abs/1906.01563}, \href
  {http://arxiv.org/abs/1906.01563} {\path{arXiv:1906.01563}}.

\bibitem{CODE}
A.~A. Hassan.
\newblock Abu-pruning-hamiltonian-nn.
\newblock \url{https://github.com/MathePhysics/Abu-pruning-Hamiltonian-NN},
  2023.

\bibitem{kadanoff2000statics}
L.~P. Kadanoff.
\newblock {\em Statics, Dynamics, and Renormalization}.
\newblock World Scientific Publishing Co Inc, 2000.

\bibitem{INVARIANT2}
A.~N. Kolmogorov.
\newblock The local structure of turbulence in incompressible viscous fluid for
  very large reynolds numbers.
\newblock {\em Proceedings: Mathematical and Physical Sciences},
  434(1890):9--13, 1991.
\newblock URL: \url{http://www.jstor.org/stable/51980}.

\bibitem{LeCun2015}
Y.~LeCun, Y.~Bengio, and G.~Hinton.
\newblock Deep learning.
\newblock {\em Nature}, 521:436--444, 2015.

\bibitem{PRUNE_FILTERS}
H.~Li, A.~Kadav, I.~Durdanovic, H.~Samet, and H.~P. Graf.
\newblock Pruning filters for efficient convnets.
\newblock {\em CoRR}, abs/1608.08710, 2016.
\newblock URL: \url{http://arxiv.org/abs/1608.08710}, \href
  {http://arxiv.org/abs/1608.08710} {\path{arXiv:1608.08710}}.

\bibitem{HENRY}
H.~W. Lin, M.~Tegmark, and D.~Rolnick.
\newblock Why does deep and cheap learning work so well?
\newblock {\em Journal of Statistical Physics}, 168(6):1223--1247, jul 2017.
\newblock URL: \url{https://doi.org/10.1007%2Fs10955-017-1836-5}, \href
  {https://doi.org/10.1007/s10955-017-1836-5}
  {\path{doi:10.1007/s10955-017-1836-5}}.

\bibitem{Lu2021}
L.~Lu, P.~Jin, and G.~E. Karniadakis.
\newblock Deeponet: Learning nonlinear operators for identifying differential
  equations based on the universal approximation theorem of operators.
\newblock {\em CoRR}, abs/1910.03193, 2019.
\newblock URL: \url{http://arxiv.org/abs/1910.03193}, \href
  {http://arxiv.org/abs/1910.03193} {\path{arXiv:1910.03193}}.

\bibitem{IMP2}
J.~Maene, M.~Li, and M.~Moens.
\newblock Towards understanding iterative magnitude pruning: Why lottery
  tickets win.
\newblock {\em CoRR}, abs/2106.06955, 2021.
\newblock URL: \url{https://arxiv.org/abs/2106.06955}, \href
  {http://arxiv.org/abs/2106.06955} {\path{arXiv:2106.06955}}.

\bibitem{marios}
M.~Mattheakis, D.~Sondak, A.~S. Dogra, and P.~Protopapas.
\newblock Hamiltonian neural networks for solving equations~of motion.
\newblock {\em Physical Review E}, 105(6), jun 2022.
\newblock URL: \url{https://doi.org/10.1103%2Fphysreve.105.065305}, \href
  {https://doi.org/10.1103/physreve.105.065305}
  {\path{doi:10.1103/physreve.105.065305}}.

\bibitem{MEHTA}
P.~Mehta and D.~J. Schwab.
\newblock An exact mapping between the variational renormalization group and
  deep learning, 2014.
\newblock \href {http://arxiv.org/abs/1410.3831} {\path{arXiv:1410.3831}}.

\bibitem{COMPUTERVISION}
A.~S. Morcos, H.~Yu, M.~Paganini, and Y.~Tian.
\newblock One ticket to win them all: generalizing lottery ticket
  initializations across datasets and optimizers, 2019.
\newblock \href {http://arxiv.org/abs/1906.02773} {\path{arXiv:1906.02773}}.

\bibitem{NO_IMP_THEORY}
M.~Paul, F.~Chen, B.~W. Larsen, J.~Frankle, S.~Ganguli, and G.~K. Dziugaite.
\newblock Unmasking the lottery ticket hypothesis: What's encoded in a winning
  ticket's mask?
\newblock In {\em The Eleventh International Conference on Learning
  Representations}, 2023.
\newblock URL: \url{https://openreview.net/forum?id=xSsW2Am-ukZ}.

\bibitem{redman}
W.~T. Redman, T.~Chen, Z.~Wang, and A.~S. Dogra.
\newblock Universality of winning tickets: A renormalization group perspective.
\newblock {\em Proceedings of the 39th International Conference on Machine
  Learning (ICML)}, 39, 2022.

\bibitem{ROSENFELD}
J.~S. Rosenfeld, J.~Frankle, M.~Carbin, and N.~Shavit.
\newblock On the predictability of pruning across scales.
\newblock {\em CoRR}, abs/2006.10621, 2020.
\newblock URL: \url{https://arxiv.org/abs/2006.10621}, \href
  {http://arxiv.org/abs/2006.10621} {\path{arXiv:2006.10621}}.

\bibitem{Sabatelli2020}
M.~Sabatelli, M.~Kestemont, and P.~Geurts.
\newblock On the transferability of winning tickets in non-natural image
  datasets.
\newblock {\em CoRR}, abs/2005.05232, 2020.
\newblock URL: \url{https://arxiv.org/abs/2005.05232}, \href
  {http://arxiv.org/abs/2005.05232} {\path{arXiv:2005.05232}}.

\bibitem{strogatz1994nonlinear}
S.~H. Strogatz.
\newblock Nonlinear dynamics and chaos: With applications to physics, biology,
  chemistry, and engineering.
\newblock {\em Studies in Nonlinearity}, 1994.

\bibitem{INVARIANT1}
K.~G. Wilson.
\newblock Renormalization group and critical phenomena. i. renormalization
  group and the kadanoff scaling picture.
\newblock {\em Phys. Rev. B}, 4:3174--3183, Nov 1971.
\newblock URL: \url{https://link.aps.org/doi/10.1103/PhysRevB.4.3174}, \href
  {https://doi.org/10.1103/PhysRevB.4.3174}
  {\path{doi:10.1103/PhysRevB.4.3174}}.

\bibitem{wilson1975renormalization}
K.~G. Wilson.
\newblock The renormalization group: Critical phenomena and the kondo problem.
\newblock {\em Rev. Mod. Phys.}, 47:773--840, Oct 1975.
\newblock URL: \url{https://link.aps.org/doi/10.1103/RevModPhys.47.773}, \href
  {https://doi.org/10.1103/RevModPhys.47.773}
  {\path{doi:10.1103/RevModPhys.47.773}}.

\bibitem{yang2023neuraltangent}
H.~Yang and Z.~Wang.
\newblock On the neural tangent kernel analysis of randomly pruned neural
  networks.
\newblock 2023.
\newblock URL: \url{https://arxiv.org/pdf/2203.14328.pdf}.

\bibitem{zhang2019alllayers}
C.~Zhang, S.~Bengio, and Y.~Singer.
\newblock Are all layers created equal?
\newblock {\em Journal of Machine Learning Research}, 2020.
\newblock URL: \url{https://www.jmlr.org/papers/volume23/20-069/20-069.pdf}.

\bibitem{zhou2019deconstructing}
H.~Zhou, J.~Lan, R.~Liu, and J.~Yosinski.
\newblock Deconstructing lottery tickets: Zeros, signs, and the supermask.
\newblock {\em CoRR}, abs/1905.01067, 2019.
\newblock URL: \url{http://arxiv.org/abs/1905.01067}, \href
  {http://arxiv.org/abs/1905.01067} {\path{arXiv:1905.01067}}.

\end{thebibliography}

\end{document}